\documentclass[journal]{IEEEtai}

\usepackage{import}
\usepackage{preamble}

\title{Gransformer: Transformer-based Graph Generation}

\author{Ahmad Khajenezhad and Seyed Ali Osia and Mahmood Karimian and Hamid Beigy 
\thanks{The authors are with the Department of Computer Engineering, Sharif University of Technology, Tehran, Iran.\protect\\ E-mails: \{khajenezhad, osia, mkarimian\}@ce.sharif.edu and beigy@sharif.edu.}
}

\begin{document}
\maketitle

\begin{abstract}
Transformers have become widely used in various tasks, such as natural language processing and machine vision. This paper proposes Gransformer, an algorithm based on Transformer for generating graphs. We modify the Transformer encoder to exploit the structural information of the given graph. The attention mechanism is adapted to consider the presence or absence of edges between each pair of nodes. We also introduce a graph-based familiarity measure between node pairs that applies to both the attention and the positional encoding. This measure of familiarity is based on message-passing algorithms and contains structural information about the graph. Also, this measure is autoregressive, which allows our model to acquire the necessary conditional probabilities in a single forward pass. In the output layer, we also use a masked autoencoder for density estimation to efficiently model the sequential generation of dependent edges connected to each node. In addition, we propose a technique to prevent the model from generating isolated nodes without connection to preceding nodes by using BFS node orderings. We evaluate this method using synthetic and real-world datasets and compare it with related ones, including recurrent models and graph convolutional networks. Experimental results show that the proposed method performs comparatively to these methods.
\end{abstract}

\section{Introduction}
\IEEEPARstart{D}{eep} graph generation models are currently used for various applications ranging from biology to drug discovery to social network analysis. Among different graph generation methods, autoregressive models have shown promising results in generating diverse graphs. These models generate graphs by sequentially generating nodes and edges based on previously generated components. 

Some autoregressive models are based on recurrent neural networks and do not use the structural properties of the graph \cite{graphRNN}, while some other methods use graph convolutional networks to directly benefit from the structural information \cite{GRAN}. There is also a prior study that substitutes recurrent neural networks with transformer networks and meanwhile tries to incorporate graph convolutional networks to use the graph structural information \cite{GRAM}. A basic challenge for autoregressive models that use graph convolutional networks for graph generation is that during their training process, they need to run a separate forward pass for each component (node, edge, or subgraph) to compute the likelihood of that component based on the preceding components. In This paper, we propose an autoregressive model based on transformers, that benefits from graph structural information by using some alternative method to the graph convolutional layers, which enables it to compute the likelihood of the input graph in a single forward pass. In the remainder of this section, we will describe prior autoregressive models, their shortcomings and an overview of our proposed model.


The first autoregressive model proposed for graph generation used graph neural networks (GNN) in a very time-consuming manner to generate graphs~\cite{DeepGMG}. You et al. proposed a method for generating larger graphs autoregressively using recurrent neural networks~\cite{graphRNN}. 
A few number of prior studies have used transformers instead of recurrent neural networks for graph generation. Fan and  Huang \cite{AGE} used a simple encoder-decoder transformer model for conditional graph generation, where the goal is to generate a graph, given another graph as input. Kawai et al. \cite{GRAM} tackled the problem of unconditional graph generation, using a modified transformer encoder, which was enriched by graph convolutional layers. 

Graph convolutional layers are common tools for extracting structural information from graphs, and  the idea of using them in autoregressive models was previously proposed in \cite{GRAN}. Since a graph convolutional layer collects the data from all neighboring nodes of each node, it violates the autoregressive property. Thus, to satisfy the autoregressive property, these models need a separate forward pass on a subset of the input graph to compute the conditional probability of each node given the preceding nodes. Therefore, they need to perform $\mathcal{O}(n)$ forward passes to compute the likelihood of an input graph with $n$ nodes. 
Our goal is to propose an alternative model for injecting structural information into a transformer-based model that retains the autoregressive property and can compute all conditional probabilities in a single forward pass for the entire graph. 
    
In this paper, we propose Gransformer, which is an autoregressive transformer-based graph generation model. Gransformer uses a single masked transformer encoder to compute all the required conditional probabilities in one forward pass.\footnote{Alternatively, we can consider our model as a single decoder of a transformer model with no cross-attention layers.}
Nodes in the network are represented by their preceding adjacency vectors, that is, the vectors that indicate their connections to the preceding nodes. There is a unit corresponding to each graph node in each self-attention layer of the transformer encoder. To determine the output of each unit, we apply a lower triangular mask to the layers of self-attention, so that only the information from the preceding nodes is taken into account. 

In addition, we use a shared Masked Autoencoder for Distribution Estimation (MADE) \cite{MADE} in the output layer to model the sequential dependency between the edge probabilities of each node to all the preceding nodes.
MADE is a well-known and efficient autoregressive model that has inspired other generative models such as~\cite{papamakarios2017masked,khajenezhad2020masked}; 
MADE helps us easily model the dependency of edge generation and efficiently generate edge probabilities all at once.

In order to integrate the structural properties of the graph to the single forward pass of a transformer encoder, we use the following two techniques: first, we change the attention formulation of the model according to the presence or absence of edges between each pair of nodes. Second, we introduce a measure of familiarity between each pair of nodes and use it to modify the attention mechanism and the positional encoding formula. This measure is derived from the message-passing algorithm and represents the structural proximity between two nodes. This measure is also autoregressive since the familiarity function of nodes $j$ and $i$ ($j \ge i$) is computed only from the subgraph that contains the first $j$ nodes of the graph.

As with some other autoregressive models, such as the models in \cite{graphRNN}, we use the BFS ordering of the nodes and then reorder the adjacency matrix based on this ordering. Therefore, in a sequence of nodes, it is not possible to have a node that is not connected to any of the preceding nodes. Due to this fact, we need to modify the predicted likelihoods of our model in a way that prevents generating nodes with no connection to the preceding ones. During the training, it can be obtained by a simple modification in the loss function. For the generation phase, we propose a sampling method for generating nodes from the corresponding probability distribution, to avoid generating nodes that are not connected to any of the preceding nodes.
The major contributions of this paper can be summarized as follows:
\begin{itemize}
    \item We propose a graph generation model with a single transformer encoder that utilizes the graph's structural properties.
    
    \item To model edge dependencies efficiently, we propose using a shared MADE in the output layer.
    
    \item With the help of the presence or absence of edges between each pair of nodes, we modify the attention mechanism of the transformer model.
    
    \item An autoregressive familiarity measure between nodes is introduced to represent the structural properties of the graphs. In both attention layers and positional encoding, we use this measure.
    
    \item The model is prevented from generating isolated nodes using a proper sampling method based on the corresponding probability distribution.
\end{itemize}
	
In this paper, we only consider generating simple undirected graphs, with no self-loops, no duplicate edges, and no node or edge labels. However, Gransformer can be easily extended to handle these cases.

It is worth noting that from a general point of view \cite{faez2021deep}, deep models for graph generation can be categorized into autoregressive models~\cite{DeepGMG,graphRNN,GRAN,BiGG, GRAM,AGE}, autoencoder-based ones~\cite{VGAE, GraphVAE, Nevae}, adversarial models~\cite{Netgan, condgen, molgan}, reinforcement learning-based methods~\cite{gcpn,graphopt}, and flow-based models~\cite{liu2019graph,madhawa2019graphnvp,shi2020graphaf}. There also exists a new track of generating graphs using diffusion models \cite{diff1,diff2,diff3,diff4,diff5}.

	
The rest of this paper is organized as follows: In Section \ref{sec:proposedMethod} we present the proposed method. First, we discuss the transformer and the basic skeleton of the model. Then we state how MADE is used for dependent edge generation,  explain our ideas for integrating graph structural properties, and discuss the strategy to avoid generating all zeros adjacency vectors. Experimental results are presented in Section \ref{sec:experiments}. Finally, Section \ref{sec:conclusion} concludes the paper.
	
\section{Related Work}
In this section, we briefly review research in graph generation. Existing models for generating deep graphs can be categorized from different points of view. Some models generate an entire graph at once \cite{VGAE,GraphVAE,MPGVAE}, while others generate a graph node by node \cite{MolMP,MolecularRNN,graphRNN}, edge by edge \cite{bacciu2019graph,bacciu2020edge,GraphGen}, or block by block \cite{GRAN,GrAD}.
As mentioned in the previous section, graph generation models can also be divided into autoencoder-based models \cite{VGAE, GraphVAE, Nevae}, adversarial models \cite{Netgan, condgen, molgan}, reinforcement learning-based methods \cite{gcpn,graphopt}, flow-based models~\cite{liu2019graph,madhawa2019graphnvp,shi2020graphaf}, autoregressive models~\cite{DeepGMG,graphRNN,GRAN,BiGG, GRAM,AGE}, and diffusion-based models~\cite{diff1,diff2,diff3,diff4,diff5}.

The models can also be classified by the type of graphs they can generate. Some models generate graphs with node or edge features~\cite{MolMP,MolecularRNN,GraphGen, GRAM }, while some others only generate graphs without any feature~\cite{graphRNN,bacciu2019graph,bacciu2020edge,GRAN}. In addition, some models may incorporate specific constraints in their construction process to generate some graphs with certain features, such as chemical rules in molecule generation~\cite{gcpn}.

Another distinguishing feature is how these models use the structural properties of the input graphs. Some models use either the graph convolutional layers \cite{VGAE,GraphVAE,GRAN} or other graph neural networks~\cite{DeepGMG}, and some other ones use random walks on the graph~\cite{Netgan,MMGAN}. The models that generate the graph sequentially also benefit from the structural information using node orders based on specific graph traversals such as breath-first search (BFS~ \cite{graphRNN,GRAN}.
Furthermore, autoencoder-based graph generators can be divided into methods that use a single graph as training data \cite{VGAE,DGVAE,Graphite} and those that use a collection of graphs as training data~\cite{GraphVAE,MPGVAE}.

It is worth noting that the graph generation problem is closely related to the node embedding problem, as many graph generation models use node embedding as an intermediate step~\cite{DEFactor,gcpn}. Furthermore, many graph generation models use similar graph processing techniques as node embedding models, such as deep node embedding models based on random walks~\cite{DeepWalk,Node2Vec,SimNet} and graph convolutional networks~\cite{GCNClassification,GraphSage}.

In the rest of this section, we briefly explain how previous autoregressive models work. Li et al. 
modeled the graph generation as a sequence of decisions~\cite{DeepGMG}. They assign a feature vector to each node and update them after each step of the decision-making sequence, using their and their neighbors' values. Decision-making is made by sampling from probability mass functions computed from these feature vectors. The sequence of decisions is as follows:
\begin{enumerate}
    \item The model decides to add a new node or finish the process. When it adds a new node, it goes to step 2.
    \item It adds a new node, and the model iteratively decides to add a new edge to this node (the last node) or finishes adding edges to this new node and returns to step 1. For adding a new edge to the last node, the model goes to step 3.
    \item The model computes a score for each node except the last node. Then, a softmax function is run on the scores to compute the probability of selecting each node. A node is selected using the corresponding probability and connected to the last node. Then, the model returns to step 2.
\end{enumerate}

Although this research \cite{DeepGMG} opened the direction of autoregressive graph generation, their functions were too time-consuming, and their method is not usable for large-scale graph generation. You et al. simplified the sequence of graph generation by the collaboration of two recurrent neural networks~\cite{graphRNN}: a graph-level RNN and an edge-level one. Each forward step of the graph-level RNN corresponds to adding a new node to the graph, and this RNN encodes the ''graph state'' vector $h_i$ in its hidden state after its $i$-th step. After each forward step of the graph-level RNN, a complete forward pass of the edge-level RNN is run. This edge-level RNN outputs the binary adjacency vector $S_i$ of length $(i-1)$ in $i-1$ steps. The $j$-th element of $S_i$ shows whether the $i$-th node is connected to the $j$-th node or not. The hidden state of the edge-level RNN is initialized by $h_i$. After running the forward pass of the edge-level RNN, $S_i$ is passed to the graph-level RNN to take its next step and compute $h_{i+1}$ using $h_i$ and $S_i$. Also, the graph-level RNN is responsible for deciding when to terminate the graph generation.

Since they do not use the structural information of the graph explicitly, an autoregressive method based on the graph convolutional neural networks presented in~\cite{GRAN}, which is faster than the method given in~\cite{DeepGMG}, and can generate large graphs block-wise. Each iteration of this method adds a new block of $B$ nodes to the current generated graph. Suppose that the sub-graph of the first $(t-1)B$ nodes is generated, and a hidden vector is assigned to each of these $(t-1)B$ nodes. This algorithm first extends this graph of $(t-1)B$ nodes to a hypothetical graph of $tB$ nodes by assuming the $t$-th block as a fully connected graph of $B$ nodes, which are all connected to all the preceding $(t-1)B$ nodes as well. It also initializes the hidden vectors on the nodes of the $t$-th block to some predefined values. Then, several layers of a spatial GCN network are run on this graph to update the hidden vectors on all the $tB$ nodes. After that, for each edge connecting to the nodes of the $t$-th block, the model decides to keep or remove that edge, based on the hidden vectors on the two ends of that edge. The remaining graph is the sub-graph of the first $tB$ nodes.

The model proposed in \cite{GRAM} substitutes the recurrent neural networks used in \cite{graphRNN} with transformers and meanwhile uses the graph convolutional networks for structural information similar to the method given in~\cite{GRAN}. They combine the features extracted from a graph convolutional layer with the attention layer of the transformer. A beneficial property of transformers is that it is easy to satisfy their autoregressive property, using a simple autoregressive mask on the attention layer. But, when combining transformers with graph convolutional layers, they cannot make them autoregressive by using a simple mask, because graph convolutional layers transfer the graph information between different nodes and violate the autoregressive property. Therefore, to calculate the adjacency vector of the $i$-th node, GRAM runs a separate forward pass on the sub-graph of the first $i$ nodes of the graph. This issue makes this method time-consuming.
Here, in this paper, we propose a new way to address this issue. In fact, we present a model to substitute the recurrent neural networks used in \cite{graphRNN} with transformers and use the structural information of the graph similarly to the method given in~\cite{GRAN} while not having the time-consuming issue of this method. 

\section{The Proposed Method} \label{sec:proposedMethod}
    
In this section, we first discuss transformers and how they are related to graph neural networks. Then, we present a method that uses a transformer for graph generation tasks. Two main challenges are the generation of dependent edges and the usage of structural information, which we will address. Finally, we discuss how to prevent the model from generating isolated nodes.
    
\subsection{Transformer and Graph Generation}

The typical form of a transformer model adopts an encoder-decoder architecture. The encoder transforms the input sequence into an embedding one. The decoder generates the output sequence in an autoregressive manner. The decoder generates the output sequence by utilizing the encoder's output and an extra input sequence. Encoders and decoders can also be used independently to solve different problems. For instance,  BERT is a transformer encoder for language understanding~\cite{bert}, while GPT-3 is a state-of-the-art natural language generative model based on a transformer decoder~\cite{gpt3}.

Transformer decoder takes two sequences, the output of the encoder and the target sequence aims to generate. Since we have only one sequence in our task that works both as the input and output of the model, we use the encoder of the transformer structure. However, we use the decoder's mask to satisfy the autoregressive property. This mask restricts the $i$-th unit in each layer to be computed just from the first $i$ units of the previous layer. Alternatively, one can consider our model as a single decoder without attention to an encoder's output.
    
The main building block of the transformer encoder is the self-attention layer that learns a weighting function to compute the weighted sum of input \textit{values} $V$. The proposed method computes the weights assigned to $V$ by using a compatibility function between \textit{queries} $Q$ and the corresponding \textit{keys} $K$~\cite{transformers}. For the $t$-th attention layer, input vectors $v_1^{(t)}, v_2^{(t)}, \ldots, v_{n}^{(t)}$ (the rows of matrix $V^{(t)}$) will be transformed to output vectors $v_1^{(t+1)}, v_2^{(t+1)}, \ldots, v_{n}^{(t+1)}$ (the rows of matrix $V^{(t+1)}$) through the following equations.
    
\begin{align}
    Att^{(t)} &= M\odot\text{softmax}\left(\frac{Q^{(t)} \times K^{(t)T}}{\sqrt{d_k}}\right),\label{eq:transformerAtt}\\
	 Q^{(t)} &= V^{(t)} \times W_Q^{(t)} + b_Q^{(t)}, \label{eq:transformerQ}\\
	 K^{(t)} &= V^{(t)} \times W_K^{(t)} + b_K^{(t)},
	   \label{eq:transformerK}\\
	 V^{(t+1)} &= Attn^{(t)} \times \left( V^{(t)} \times W_V^{(t)} + b_V^{(t)}\right), \label{eq:transformerV}
\end{align}
where $Attn^{(t)}$ is the attention of the units in the $(t+1)$-th layer to the units in the $t$-th layer. $W_Q^{(t)}$, $W_K^{(t)}$, $W_V^{(t)}$, $b_V^{(t)}$, $b_Q^{(t)}$ and $b_K^{(t)}$ are trainable parameters of the model. $M$ is the mask matrix, and $\odot$ denotes the element-wise multiplication.
	
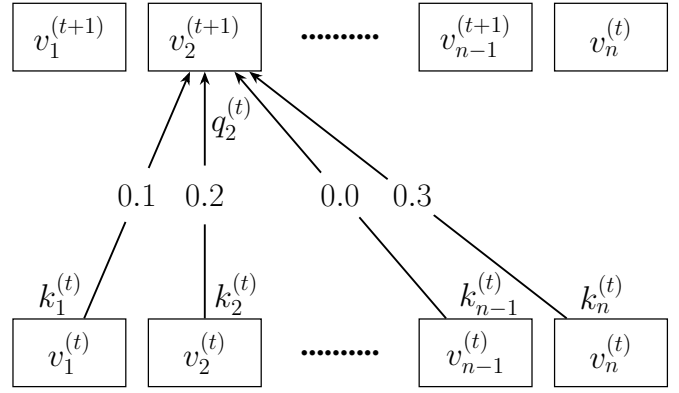
\begin{figure}[!ht]
    \begin{center}	
        \scalebox{0.6}{\def \sc {0.8}
\begin{tikzpicture}[scale=\sc]
\def \fs {\large}

\begin{scope}[every node/.style={thick,draw, minimum width=2.5cm, minimum height=1.5cm, font=\fs}]
    \node (D1) at (0,0) {$v_1^{(t)}$};
    \node (D2) at (3,0) {$v_2^{(t)}$};
    \node (D3) at (9,0) {$v_{n-1}^{(t)}$};
    \node (D4) at (12,0) {$v_n^{(t)}$};
    \node (U1) at (0,7) {$v_1^{(t+1)}$};
    \node (U2) at (3,7) {$v_2^{(t+1)}$};
    \node (U3) at (9,7) {$v_{n-1}^{(t+1)}$};
    \node (U4) at (12,7) {$v_n^{(t)}$};
\end{scope}

\begin{scope}[every node/.style={font=\fs}]
    \node (Ddots) at (6,0) {\textbf{..........}};
    \node (Udots) at (6,7) {\textbf{..........}};
\end{scope}


\begin{scope}[>={Stealth[black]},
              every edge/.style={draw=black,very thick}]
\draw [->] (D1) edge node {} (U2);
\path [->] (D2) edge node {} (U2);
\path [->] (D3) edge node {} (U2);
\path [->] (D4) edge node {} (U2);
\end{scope}

\begin{scope}[every node/.style={thick,circle, fill=white, font=\fs}]
    \node (Att1) at (1.5,3.5) {$0.1$};
    \node (Att2) at (3,3.5) {$0.2$};
    \node (Att3) at (6,3.5) {$0.0$};
    \node (Att4) at (7.6,3.5) {$0.3$};
\end{scope}

\begin{scope}[every node/.style={thick, font=\fs}]
    \node (K1) at (-0.2, 1.3) {$k_1^{(t)}$};
    \node (K2) at (3.7, 1.3) {$k_2^{(t)}$};
    \node (K3) at (9.3, 1.3) {$k_{n-1}^{(t)}$};
    \node (K4) at (11.8, 1.3) {$k_n^{(t)}$};
    \node (Q2) at (3.6, 5.2) {$q_2^{(t)}$};
\end{scope}

\end{tikzpicture}}
    \end{center}
    \caption{The schema of the self-attention layer of the transformer encoder. As illustrated in this figure, for each $1 \le i \le n$, the value $v_i^{(t+1)}$ is computed from values $v_j^{(t)}$ for all $1 \le j \le n$. From this point of view, this layer is similar to a graph convolutional layer on a fully connected graph. When we are using an autoregressive mask, $v_i^{(t+1)}$ would be computed from values $v_j^{(t)}$ for all $1 \le j \le i$, and this operation would be similar to a graph convolutional layer on a fully connected directed acyclic graph.}
    \label{fig:transformerAttention}
\end{figure}

Based on the graph convolutional networks paradigm, the Transformer encoder is similar to an attention-based spatial graph convolutional network on a fully connected graph. Let us map the $n$ units in each self-attention layer of the transformer encoder to the $n$ nodes of a hypothetical graph. In each self-attention layer, the attention of each unit to another unit assumes a hypothetical edge between the corresponding pair of nodes in the graph. Since without considering a mask, every unit in the encoder has attention to all the units, this structure is similar to the graph convolution on a fully connected graph i.e., there is a directed edge from each node to each other node. Using the decoder's mask, this structure would be similar to graph convolution on a directed graph in which there are edges from each node to all subsequent nodes (see figure \ref{fig:transformerAttention}). We will use this GCN-based viewpoint to the transformers in section \ref{sec:typedEdges} to use graph structural information in the process of graph generation.

Let us now discuss how to use the Transformer encoder to generate graphs. Our autoregressive model generates a graph node-by-node and edge-by-edge in a sequential manner. Given an incomplete graph of $k-1$ nodes, the proposed model must be able to decide about the $k$-th node. It has to estimate the probability of connecting the $k$-th node to any of the preceding $k-1$ nodes. Since we use a node ordering $\pi$ of nodes for the sequential graph generation, let us reformulate the log-likelihood of a graph $G$ in the following form:

\begin{small}
\begin{align}
    \log P(G) &= \log P(n)\displaystyle\sum_{\pi}P(\pi)P(G \vert \pi) \nonumber \\
    &= \log P(n) + \log \displaystyle\sum_{\pi}P(\pi)P(G \vert \pi)
\end{align}
\end{small}

We estimate the discrete distribution $P(n)$ by maximum likelihood estimation using a simple empirical estimation. It remains to train our model parameters to optimize $\log \displaystyle\sum_{\pi}P(\pi)P(G \vert \pi)$. Since we use a uniform distribution over all BFS orders, $P(\pi)$ can be considered as a constant value and the only remaining objective function will be $\log \displaystyle\sum_{\pi}P(G \vert \pi)$. Similar to most prior auto-regressive works \cite{DeepGMG,graphRNN,GRAM} we optimize the surrogate objective function $\log \displaystyle\sum_{\pi}P(G \vert \pi)$. Similar to prior words, in the training process, for each input graph in each mini-batch, we select a single random BFS-based ordering of its nodes.

Assuming any order $\pi$ of nodes, we use the probability chain rule for graph generation and graph density estimation.
Let $A^{\pi}$ be the adjacency matrix of the graph of size $n$ under the ordering $\pi$. Let $L^{\pi}$ be the lower triangular form of $A^{\pi}$. Let $L_i^{\pi}$ also be the $i$-th row of $L^{\pi}$. We encode a graph with the sequence of vectors $L_1^{\pi}, L_2^{\pi}, \ldots, L_n^{\pi}$. We need a model that can generate $L_{i+1}^{\pi}$ using $L_1^{\pi}, L_2^{\pi}, \ldots, L_{i}^{\pi}$. In other words, by following the probability chain rule, we will find that:

\begin{small}
\begin{align} \label{eq:chainRule}
        P(G \vert \pi) &= P(A^{\pi} \vert \pi) = \prod_i P(L_i^{\pi} \vert  L_1^{\pi}, \ldots, L_{i-1}^{\pi})
\end{align}
\end{small}

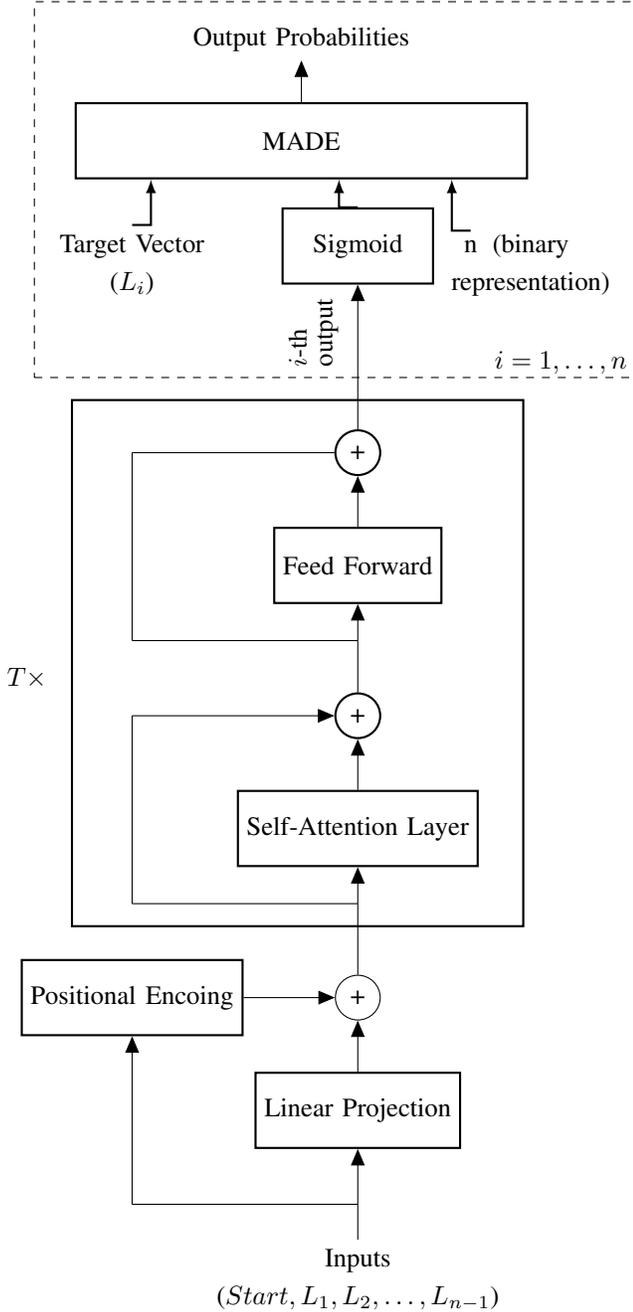
\begin{figure}[!ht]
 \centering
  \scalebox{1}{\def \sc {0.8}
\begin{tikzpicture}[scale=\sc]

\def \Ox {0}
\def \Oy {0}
\def \A {A}

\node (input\A{}L) at (5+\Ox,-0.25+\Oy) [minimum width=\sc*2cm,minimum height=\sc*0.5cm] {($Start, L_1, L_2, \ldots, L_{n-1}$)};

\node (input\A) at (5+\Ox,0.25+\Oy) [minimum width=\sc*2cm,minimum height=\sc*0.5cm] {Inputs};

\node (cp0\A) at (5+\Ox,1+\Oy) [] {};

\path [-] (input\A) edge node {} (cp0\A.center);

\node (linProject\A) at (5+\Ox,2.25+\Oy) [draw,thick,minimum width=\sc*2cm,minimum height=\sc*1cm] {Linear Projection};

\path [->] (cp0\A.center) edge node {} (linProject\A);

\node (cp01\A) at (2+\Ox,1+\Oy) [] {};

\path [-] (cp0\A.center) edge node {} (cp01\A.center);

\node (posEnc\A) at (2+\Ox,3.75+\Oy) [draw,thick,minimum width=\sc*2cm,minimum height=\sc*1cm] {Positional Encoing};

\path [->] (cp01\A.center) edge node {} (posEnc\A);

\node (add1\A) at (5+\Ox,3.75+\Oy) [circle,draw,minimum size=\sc*0.4cm] {+};

\path [->] (posEnc\A) edge node {} (add1\A);

\path [->] (linProject\A) edge node {} (add1\A);

\node (block\A) at (4.2+\Ox,8.2+\Oy) [draw,thick,minimum width=\sc*6cm,minimum height=\sc*7cm] {};

\node (nTimes\A) at (0.6+\Ox,8+\Oy) [] {$T\times$};

\node (cp1\A) at (5+\Ox,5+\Oy) [] {};

\node (cp2\A) at (2+\Ox,5+\Oy) [] {};

\node (cp3\A) at (2+\Ox,7.5+\Oy) [] {};

\path [-] (add1\A) edge node {} (cp1\A.center);

\path [-] (cp1\A.center) edge node {} (cp2\A.center);

\path [-] (cp2\A.center) edge node {} (cp3\A.center);

\node (attention\A) at (5+\Ox,6+\Oy) [draw,thick,minimum width=\sc*2cm,minimum height=\sc*1cm] {Self-Attention Layer};

\path [->] (cp1\A.center) edge node {} (attention\A);

\node (add2\A) at (5+\Ox,7.5+\Oy) [circle,draw,thick,minimum size=\sc*0.4cm] {+};

\path [->] (attention\A) edge node {} (add2\A);

\path [->] (cp3\A.center) edge node {} (add2\A);

\node (cp4\A) at (5+\Ox,8.5+\Oy) [] {};

\node (feedforward\A) at (5+\Ox,9.5+\Oy) [draw,thick,minimum width=\sc*2cm,minimum height=\sc*1cm] {Feed Forward};

\path [-] (add2\A) edge node {} (cp4\A.center);

\path [->] (cp4\A.center) edge node {} (feedforward\A);

\node (cp5\A) at (2+\Ox,8.5+\Oy) [] {};

\path [-] (cp4\A.center) edge node {} (cp5\A.center);

\node (add3\A) at (5+\Ox,11+\Oy) [circle,draw,thick,minimum size=\sc*0.4cm] {+};

\path [->] (feedforward\A) edge node {} (add3\A);

\node (cp6\A) at (2+\Ox,11+\Oy) [] {};

\path [-] (cp5\A.center) edge node {} (cp6\A.center);

\path [-] (cp6\A.center) edge node {} (add3\A);

\def \OH {1}

\node (sigmoid\A) at (5+\Ox, \OH + 12.75+\Oy) [draw,thick,minimum width=\sc*2cm,minimum height=\sc*1cm] {Sigmoid};

\node (MADE\A) at (4.25+\Ox, \OH + 14.15+\Oy) [draw,thick,minimum width=\sc*6cm,minimum height=\sc*1cm] {MADE};

\node (n\A) at (6.5+\Ox, \OH + 12.75+\Oy) [] {$n$};
\node (n\A{}b) at (7.3+\Ox, \OH + 12.75+\Oy) [] {{\small (binary}};
\node (n\A{}r) at (7.3+\Ox, \OH + 12.25+\Oy) [] {{\small representation)}};

\node (target\A) at (2+\Ox, \OH + 12.75+\Oy) [] {Target Vector};
\node (target\A{}L) at (2+\Ox, \OH + 12.25+\Oy) [] {($L_i$)};

\path [->] (add3\A) edge node {} (sigmoid\A);

\draw[-latex, thick] (sigmoid\A.north) -| ([xshift=0.5cm]MADE\A.south);
\draw[-latex, thick] (target\A.north) -| ([xshift=-2cm]MADE\A.south);
\draw[-latex, thick] (n\A.north) -| ([xshift=2cm]MADE\A.south);

\node (output\A) at (4.25+\Ox, \OH + 15.5+\Oy) [] {Output Probabilities};
\path [->] (MADE\A) edge node {} (output\A);

\node[draw] (replicateMADE) at (4.7+\Ox, \OH + 13.5+\Oy) [draw,dashed,minimum width=\sc*8cm,minimum height=\sc*5cm] {};

\node (repMADEexpl) at (7.7+\Ox, 12.2+\Oy) {$i=1,\ldots,n$};

\node (sigout1) at (4.2+\Ox, 12.4+\Oy) {\rotatebox[origin=c]{90}{$i$-th}};
\node (sigout) at (4.6+\Ox, 12.6+\Oy) {\rotatebox[origin=c]{90}{output}};

\end{tikzpicture}}
  \caption{Schema of the Gransformer model. Notice that the autoregressive (MADE) network used at the output layer is shared between all the $n$ output vectors of the transformer encoder.}
\label{fig:gg-base-model}
\end{figure}

Therefore, we need the transformer model to estimate every one of these $n$ conditional probabilities. The sequence of the input vectors to the transformer encoder would be $Start, L_1^{\pi}, L_2^{\pi}, \ldots, L_n^{\pi}$ and its output vectors would be $y_1^{\pi}, y_2^{\pi}, \ldots, y_n^{\pi}, End$, where $Start$ and $End$ are the padding vectors. The $i$-th element of $y_k$ ($i < k$) determines the probability of putting an edge between the $k$-th node and the $i$-th node. Therefore, since $L_1^{\pi}, \ldots, L_n^{\pi}$ are binary vectors, the target sequence of the vectors in the output layer is $L_1^{\pi}, L_2^{\pi}, \ldots, L_n^{\pi}, End$. We use a lower triangular matrix as the mask matrix i.e., $M$ in Equation \eqref{eq:transformerAtt}, to guarantee that $y_k$ is computed only from the first $k$ inputs i.e., from $L_1^{\pi},\ldots, L_{k-1}^{\pi}$. Figure \ref{fig:gg-base-model} shows the schema of the proposed model. The model described so far is similar to the encoder part of the transformer network. However, it uses the decoder's mask, so that unit $i$ in each layer would be allowed to have attended only to the first $i$ units of the previous layer. In the following subsection, we describe the proposed extensions for the task of graph generation.
    
\subsection{Dependent Edge Generation} \label{sec:MADE}

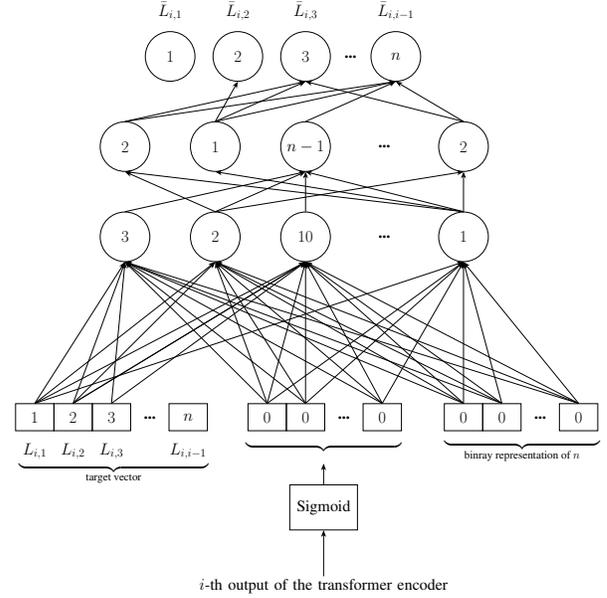
\begin{figure}[!ht]
    \centering
    \scalebox{0.3}{\begin{tikzpicture}

\def \fs {\huge}
\def \OU {8}
\def \OH {4}

\begin{scope}[every node/.style={thick,draw, minimum width=1.7cm, minimum height=1.2cm, font=\fs}]
    \node (D1) at (0,0) {$1$};
    \node (D2) at (1.7,0) {$2$};
    \node (D3) at (3.4,0) {$3$};
    \node (D4) at (6.8,0) {$n$};
    
    \node (S1) at (10.3,0) {$0$};
    \node (S2) at (12,0) {$0$};
    \node (S3) at (15.4,0) {$0$};
    
    \node (N1) at (19,0) {$0$};
    \node (N2) at (20.7,0) {$0$};
    \node (N3) at (24.1,0) {$0$};
    
\end{scope}

\node[thick,draw, minimum width=3cm, minimum height=2cm, font=\fs] (Sigmoid) at (12.8, -4) {Sigmoid};

\begin{scope}[every node/.style={thick, font=\fs}]
    \node (DL1) at (0,-1.5) {$L_{i,1}$};
    \node (DL2) at (1.7,-1.5) {$L_{i,2}$};
    \node (DL3) at (3.4,-1.5) {$L_{i,3}$};
    \node (DL4) at (6.8,-1.5) {$L_{i,i-1}$};
    
    \node (UL1) at (6 + 0, \OH + 14) {$\bar{L}_{i,1}$};
    \node (UL2) at (6 + 3, \OH + 14) {$\bar{L}_{i,2}$};
    \node (Ul3) at (6 + 6, \OH + 14) {$\bar{L}_{i,3}$};
    \node (UL4) at (6 + 10, \OH + 14) {$\bar{L}_{i,i-1}$};
    
    \node (SL) at (12.8, -1.5) {$\underbrace{\textcolor{white}{.}\hspace{6.5cm}\textcolor{white}{.}}_{}$};
    
    \node (NL) at (21.6, -1.5) {$\underbrace{\textcolor{white}{.}\hspace{6.5cm}\textcolor{white}{.}}_{\text{binray representation of }n}$};
    
    \node (SL2) at (3.5, -2.5) {$\underbrace{\textcolor{white}{.}\hspace{8cm}\textcolor{white}{.}}_{\text{target vector}}$};
    
    \node (SL2) at (12.8, -7.5) {$i$-th output of the transformer encoder};
\end{scope}

\begin{scope}[every node/.style={font=\fs}]
    \node (Ddots) at (5.1,0) {\textbf{...}};
    \node (Sdots) at (13.7,0) {\textbf{...}};
    \node (Ndots) at (22.4,0) {\textbf{...}};
    \node (H1dots) at (6 + 9.5, \OH + 4) {\textbf{...}};
    \node (H2dots) at (6 + 9.5, \OH + 8) {\textbf{...}};
    \node (Udots) at (6 + 8, \OH + 12) {\textbf{...}};
\end{scope}

\begin{scope}[every node/.style={circle,thick,draw, minimum size=2.2cm, font=\fs}]
    \node (H11) at (4 + 0, \OH + 4) {$3$};
    \node (H12) at (4 + 4, \OH + 4) {$2$};
    \node (H13) at (4 + 8, \OH + 4) {$10$};
    \node (H14) at (4 + 15, \OH + 4) {$1$};
    
    \node (H21) at (4 + 0, \OH + 8) {$2$};
    \node (H22) at (4 + 4, \OH + 8) {$1$};
    \node (H23) at (4 + 8, \OH + 8) {$n-1$};
    \node (H24) at (4 + 15, \OH + 8) {$2$};
    
    \node (U1) at (6 + 0, \OH + 12) {$1$};
    \node (U2) at (6 + 3, \OH + 12) {$2$};
    \node (U3) at (6 + 6, \OH + 12) {$3$};
    \node (U4) at (6 + 10, \OH + 12) {$n$};
\end{scope}

\begin{scope}[>={Stealth[black]},
              every edge/.style={draw=black,very thick}]
\draw [->] (SL2) edge node {} (Sigmoid);
\path [->] (Sigmoid) edge node {} (SL);

\path [->] (D1.north) edge node {} (H11.south);
\path [->] (D1.north) edge node {} (H12.south);
\path [->] (D1.north) edge node {} (H13.south);
\path [->] (D1.north) edge node {} (H14.south);
\path [->] (D2.north) edge node {} (H11.south);
\path [->] (D2.north) edge node {} (H12.south);
\path [->] (D2.north) edge node {} (H13.south);
\path [->] (D3.north) edge node {} (H11.south);
\path [->] (D3.north) edge node {} (H13.south);
\path [->] (D2.north) edge node {} (H13.south);

\path [->] (S1.north) edge node {} (H11.south);
\path [->] (S1.north) edge node {} (H12.south);
\path [->] (S1.north) edge node {} (H13.south);
\path [->] (S1.north) edge node {} (H14.south);
\path [->] (S2.north) edge node {} (H11.south);
\path [->] (S2.north) edge node {} (H12.south);
\path [->] (S2.north) edge node {} (H13.south);
\path [->] (S2.north) edge node {} (H14.south);
\path [->] (S3.north) edge node {} (H11.south);
\path [->] (S3.north) edge node {} (H12.south);
\path [->] (S3.north) edge node {} (H13.south);
\path [->] (S3.north) edge node {} (H14.south);

\path [->] (N1.north) edge node {} (H11.south);
\path [->] (N1.north) edge node {} (H12.south);
\path [->] (N1.north) edge node {} (H13.south);
\path [->] (N1.north) edge node {} (H14.south);
\path [->] (N2.north) edge node {} (H11.south);
\path [->] (N2.north) edge node {} (H12.south);
\path [->] (N2.north) edge node {} (H13.south);
\path [->] (N2.north) edge node {} (H14.south);
\path [->] (N3.north) edge node {} (H11.south);
\path [->] (N3.north) edge node {} (H12.south);
\path [->] (N3.north) edge node {} (H13.south);
\path [->] (N3.north) edge node {} (H14.south);

\path [->] (H11.north) edge node {} (H23.south);
\path [->] (H12.north) edge node {} (H21.south);
\path [->] (H12.north) edge node {} (H23.south);
\path [->] (H12.north) edge node {} (H24.south);
\path [->] (H13.north) edge node {} (H23.south);
\path [->] (H14.north) edge node {} (H21.south);
\path [->] (H14.north) edge node {} (H22.south);
\path [->] (H14.north) edge node {} (H23.south);
\path [->] (H14.north) edge node {} (H24.south);

\path [->] (H21.north) edge node {} (U3.south);
\path [->] (H21.north) edge node {} (U4.south);
\path [->] (H22.north) edge node {} (U2.south);
\path [->] (H22.north) edge node {} (U3.south);
\path [->] (H22.north) edge node {} (U4.south);
\path [->] (H23.north) edge node {} (U4.south);
\path [->] (H24.north) edge node {} (U4.south);
\path [->] (H24.north) edge node {} (U3.south);

\end{scope}

\end{tikzpicture}}
    \caption{Here is an example of a MADE network appearing at the output layer. Numbers written on the network units and inputs are the tags. Masked connections have been removed in this figure. The $i$-th output of the transformer encoder contains the information from the $i-1$ preceding nodes of the graph. Therefore, it is tagged by zero to be used by all the output units. The target vector contains the information on the edges connecting the $i$-th node to the preceding nodes. The network masks guarantee that the $j$-th output unit uses only the first $j-1$ elements of the target vector. It is worth noting that only the first $i-1$ elements of $\bar{L}_i$ are of our interest, and the rest will be treated as zero. To keep the figure simple and sparse, we ignored drawing direct links from the input layer to the output layer.}
\label{fig:MADE}
\end{figure}

Suppose that we only use a masked transformer encoder, without using any extra layers at its output layer. Then, the output sequence of the vectors from the transformer encoder will act as the probability estimation of the adjacency vectors. In other words, the first  $i-1$ elements of the $i$-th output vector would represent the probability of the presence of the edges between node $i$ and nodes $1,2,\ldots,i-1$. All these $i-1$ values are computed from the first $i-1$ input vectors. Therefore, the resulting probability for connecting node $i$ to node $j$ ($j<i$Suppose we use only a masked transformer encoder without using additional layers on the output layer. Then the output sequence of vectors from the transformer encoder serves as the probability estimate for the adjacency vectors. In other words, the first $(i-1)$ elements of the $i$-th output vector would represent the probability of the presence of edges between node $i$ and nodes $1,2,\ldots, (i-1)$. All of these $(i-1)$ values are calculated from the first $(i-1)$ input vectors. Therefore, the resulting probability of connecting node $i$ to node $j$ ($j$) depends only on the edges such as $\langle t, s \rangle$ that $t,s < i$. However, it is obvious that knowing the existence or absence of the edges $\langle 1, i \rangle$, $\langle 2, i \rangle$, $\langle (j-1), i \rangle$ has a significant impact on the decision about the existence of the edge $\langle j, i \rangle$. Therefore, we can extend equation \eqref{eq:chainRule} to  capture the sequential generation of edges:

\begin{small}
\begin{align}
    P(G \vert \pi) &= \prod_{i=1}^n \prod_{j=1:i-1} P(L_{ij}^{\pi} \vert L_1^{\pi}, \ldots, L_{i-1}^{\pi}, L_{i,1}^{\pi}, \ldots, L_{i,j-1}^{\pi})
\end{align}
\end{small}

To model this dependency, You et al. used an additional recurrent neural network in their model for sequential edge generation, which means additional training time~\cite{graphRNN}. Also, Liao et al. proposed using a mixture to capture the correlation between edges~\cite{GRAN}. However, this only increases the ability of their method to model more complex edge distributions. However, it does not fully model the dependence between edges of individual nodes (or blocks in their proposed approach).

We propose to use the Masked Autoencoder for Density Estimation (MADE) \cite{ MADE } for sequential edge generation. MADE is a fully connected feed-forward autoencoder where each unit has a tag in $\lbrace 1,2,\ldots,n \rbrace$. Also, the $n$ units in the input layer and the $n$ units in the output layer have labels $1,2,\ldots,n$. Connection from a node with label $i$ to another node labeled $j$ in the following layer will be masked if $i > j$. All paths from input nodes that have labels $j+1, j+2, \ldots, n$ to nodes with labels $j$ in the following layers will be blocked using this mask set. In the same way, it can be shown that the label of each node determines the subset of input units that are allowed to have a path to that node. Therefore, this masking strategy satisfies the autoregressive property.
  
We use a single MADE with shared parameters for generating the edges of all nodes. In the original MADE network, the input is exactly the target vector, which is supposed to be estimated in the output layer. Here, the target vector for the $i$-th node is the adjacency vector of the node $i$ that we want to see in the output layer of MADE. Therefore, for node $i$, we feed $L_i$ into the MADE network, and it tries to estimate $L_i$ in the output. But, as shown in figure \ref{fig:gg-base-model}, in addition to the target vector, we feed two more inputs to the MADE network: $n$, which is the number of nodes and the $i$-th output vector of the transformer encoder. Additional information from the other two inputs can be incorporated fully into the reconstruction of the target vector in the output layer, so they are tagged as $0$ in the input layer, despite the target vector in the input layer which should have tags from 1 to $n$. The experimental results show that adding direct links from the input units of MADE to the output units with larger tags has a significant impact on the results. Figure \ref{fig:MADE} illustrates how we use the MADE network to generate the $i$-th output vector.
    
\subsection{Using Graph Structural Properties} \label{sec:structuralInfo}

As mentioned earlier, the Gransformer is based on the Transformer, which was originally proposed for natural language processing.
In NLP tasks, the input text to the Transformer model is represented by a sequence of one-hot vectors. Similarly, we represent an input graph for our model by a sequence of vectors, $ L_1^{\pi}, L_2^{\pi}, \ldots, L_n^{\pi}$. However, this representation does not explicitly specify the structural information of the graph that can be used during the training and generation process.

In prior work, graph convolutional networks have been used for this purpose \cite{DeepGMG, GRAM, GRAN}. To infer a new node (or a new block of nodes) and its connecting edges to the previous nodes, these models consider the subgraph of the existing nodes and apply multiple graph convolutional layers to it. Therefore, $\mathcal{O}(n)$ forward passes are needed to compute the loss function. However, we intend to combine the structural graph information with our transformer-based model such that the loss function can be calculated in a single forward pass.
In Sections \ref{sec:typedEdges} and \ref{sec:gatt} we will discuss two techniques we have proposed for this goal.

\subsubsection{Using Presence/Absence of Edges} \label{sec:typedEdges}
	
In a regular transformer encoder, each node can attend to all the nodes in the previous layer. As with the transformer decoder, we restrict it to have attention to the preceding nodes using autoregressive masks. But even in this way, we do not use the structural information of the graph and ignore the edges.
	
Using different sets of key/query/value parameters for pairs of nodes with an edge between them and pairs of nodes without an edge between them allows us to import the structural information of the graph into our proposed model. To be more specific, we can rewrite the Equations \eqref{eq:transformerAtt}, \eqref{eq:transformerQ}, \eqref{eq:transformerK} and \eqref{eq:transformerV} in the following form:

\begin{small}
\begin{align}
    Att_1^{(t)} &= M \odot \text{softmax}\left(\frac{Q_1^{(t)} \times K_1^{(t)T}}{\sqrt{d_k}}\right), \\
    Att_2^{(t)} &= M \odot \text{softmax}\left(\frac{Q_2^{(t)} \times K_2^{(t)T}}{\sqrt{d_k}}\right), \\
	Q_1^{(t)} &= V_1^{(t)} \times W_{Q1}^{(t)} + b_{Q1}^{(t)}, \\
	Q_2^{(t)} &= V_2^{(t)} \times W_{Q2}^{(t)} + b_{Q2}^{(t)}, \\
	K_1^{(t)} &= V_1^{(t)} \times W_{K1}^{(t)} + b_{K1}^{(t)}, \\
	K_2^{(t)} &=  V_2^{(t)} \times W_{K2}^{(t)} + b_{K2}^{(t)}, \\
	V_1^{(t+1)} &= Att_1^{(t)} \times ( V^{(t)} \times W_{V1}^{(t)} + b_{V1}^{(t)} ),\\
	V_2^{(t+1)} &= Att_2^{(t)} \times ( V^{(t)} \times W_{V2}^{(t)} + b_{V2}^{(t)} ),\\
	V^{(t+1)}(i,j) &= \begin{cases}
	   V_1^{(t+1)} & if A(i,j)=1 \\
	   V_2^{(t+1)} & if A(i,j)=0
	\end{cases}
\end{align}
\end{small}

This concept duplicates the computation time without changing the order of time complexity. We will see in Section \ref{sec:ablation} that the idea of using presence/absence of edges in the attention layer as described above does not necessarily improve our model in practice. Actually, the model needs to collect the information from a longer distance neighborhood for each node, to benefit from the graph structural information. However, we will propose our second idea for injecting graph structural information into our model in Section \ref{sec:gatt}. Our experiments in Section \ref{sec:ablation} confirms improvement of our model using the latter method.
	
\subsubsection{Autoregressive Graph-based Familiarities} \label{sec:gatt}

This section aims to define a familiarity function $K$ with desired properties between pairs of nodes. To be more precise, $K$ would be
\begin{enumerate}
    \item based on the graph structure, or represent the graph structural information,
    \item autoregressive; i.e., if $i>j$ then $K(i,j)$ is a constant value and if $i \le j$ then $K(i,j)$ is computed only from the subgraph of the first $j$ nodes.
    \item efficiently computable for all the node pairs.
\end{enumerate}
We define $K$ as follows:

\begin{small}
\begin{align}
        K(i,j) = sigmoid(f( &g_0(i,j), \ldots, g_{n_k}(i,j), \nonumber \\
        &h_{n_k}(i,j), \ldots, h_0(i,j))) \label{eq:graph-kernel}
\end{align}
\begin{align}
g_k(i,j) &= m_k(i,j) / \max(1, \displaystyle \sum_t m_k(t,j))\label{eq:normalized-num-paths-1}\\
h_k(i,j) &= m_k(i,j) / \max(1, s_k(i,j)) \label{eq:normalized-num-paths-2}
\end{align}
\end{small}


%
where $f(.)$ is a two-layers MLP with a ReLU activation function in the first layer and a linear activation function in the second layer. The scalar variable $m_k(i,j)$ denotes the number of paths of length $k$ from node $i$ to node $j$ in the subgraph containing nodes $1$ to $j$. Also $s_k(i,j)$ denotes the number of paths of length $k$ starting from node $i$ (to any node) in the subgraph containing nodes $1$ to $j$. Here, a path means any sequence of nodes in which every two consecutive nodes are connected (so a node can appear more than once in a path). 
It is clear that if $j<i$, then according to the definition, $m_k(i,j) = 0$ for all values of $k$, and consequently $K(i,j)$ will be a constant value, independent of $i$ and $j$. 


Here we briefly justify why the number of paths of length $k$ between pairs of nodes is an appropriate value to represent the structural information of the graph. We consider a process of message passing on a subgraph of the first $j$ nodes in which each node passes its value to all its neighbors at each step.
Following this algorithm, the initial message on node $i$ is received exactly $m_k(i,j)$ times by the $j$-th node in step $k$. It is worth noting that most graph convolutional layers and graph kernels based on random walks can be restated 
by special forms of message passing. For example, in spatial graph convolutional layers, the vector at each node is computed by aggregating the vectors (or equivalent messages) from its neighboring nodes, followed by a nonlinear activation function.

In fact $g_k(i,j)$ and $h_k(i,j)$ are two different normalizations of $m_k(i,j)$. On one hand, these normalizations bound the input values to network $f$; otherwise, it may face numerical overflows for its parameter values during the training process. On the other hand, these normalizations have some probabilistic meaning from a random walk-based point of view. In other words, if we consider the subgraph of the first $j$ nodes of an undirected graph, then $g_k(i,j)$ is the probability of stopping at node $i$ if we start a random walk of length $k$ from node $j$. Also $h_k(i,j)$ is the probability of stopping at node $j$ if we start a random walk of length $k$ from node $i$.
    
We now explain how to calculate $m_0(.,.), \ldots, m_{n_k}(.,.)$ efficiently. We define $M_k$ as an upper triangular matrix whose element in the $i$-th row and $j$-th column is equal to $m_k(i,j)$. $M_0$ is an identity matrix. Let $A$ be the adjacency matrix of the graph and $A(i,j)$ be the element in row $i$ and column $j$. Assume that we have computed the function $m_{k-1}(.,.)$ or equivalently $M_{k-1}$. Each path of length $k$ from $i$ to $j$ that uses only the first $j$ nodes constitutes a first step from node $i$ to a node $t \in \lbrace 1, 2, \ldots, j \rbrace$ and a path of length $k-1$ from node $t$ to node $j$ that uses only the first $j$ nodes. Therefore for $i \le j$ we have:

\begin{small}
\begin{align}
    m_k(i,j) &= \sum_{1 \le t \le j} A(i,t) \times m_{k-1}(t,j) \nonumber \\
    &= \sum_{1 \le t \le n} A(i,t) \times m_{k-1}(t,j) \nonumber \\
    &= \left[A \times M_{k-1}\right](i,j) \hspace{0.5cm} \text{if} \hspace{0.2cm} i \le j
\end{align}
\end{small}

The second line of the above equation is because, for $t>j$, we have $m_{k-1}(t,j) = 0$. Similarly, for $m_k$ we must have $i > j \Rightarrow m_k(i,j) = 0$. Therefore
    
\begin{small}
\begin{align}
    M_k &= tri\_up(A \times M_{k-1}),
\end{align}
\end{small}
where the operator $tri\_up$ converts its input matrix to an upper triangular matrix.
    
Computing this measure for all node pairs needs $k$ multiplications of $n \times n$ matrices, which can theoretically be run either in 
$\mathcal{O}(kn^{2.3728596})$ \cite{matrixMult}, but practically in $\mathcal{O}(kn^{2.8074})$ \cite{strassen1969gaussian}
or $\mathcal{O}(kne)$ for sparse graphs, where $e$ is the number of edges (because one of the matrices in each multiplication is the adjacency matrix). In addition, $n_k$ is an integer parameter in the familiarity measure, which we set to $16$ in our experiments.

We can compute $s_0(.,.), \ldots, s_{n_k}(.,.)$ in a very similar way to $m_0(.,.), \ldots, m_{n_l}(.,.)$. If we define $S_k$ as an upper triangular matrix whose elements in the $i$-th row and $j$-th column is equal to $s_k(i,j)$, then $S_0$ will be a upper triangular all one matrix. Then using a similar discussion as above we can see that:
\begin{small}
\begin{align}
    S_k &= tri\_up(A \times S_{k-1}),
\end{align}
\end{small}
    
Finally, we will indicate the utilization of the familiarity measure \eqref{eq:graph-kernel} in our model. We multiply the familiarity function by the attention in each layer. 
In other words, we modify the attention defined in Equation \eqref{eq:transformerAtt} in the following form:

\begin{small}
\begin{align}
    Att^{(t)} &\leftarrow Att^{(t)} \odot \mathcal{K}^T,
\end{align}
\end{small}
where $\mathcal{K}$ is an $n \times n$ matrix containing $K(i,j)$ for all values of $i$ and $j$.
It is worth noting that the familiarity function defined in Equation \eqref{eq:graph-kernel} contains a two-layer neural network $f$. This neural network can have different parameter values in different layers of our model. These parameters will be learned with other network parameters through the training process.
    
Similarly, we propose a positional encoding for the nodes based on $m_k(.,.)$ for $k \in \lbrace 0, 2, \ldots, n_k \rbrace$. The intuition is that if we put a one-hot vector on each vector representing its index and let these vectors propagate over the subgraph of the first $j$ nodes for $k$ times, then the $i$-th element of the final vector (after $k$ steps of message passing) on the $j$-th node will be $m_k(i,j)$. So we define the positional encoding of node $j$ as follows:

\begin{small}
\begin{align}
    pos\_enc(j) = &f_1\left(z_0(j), z_1(j), \ldots, z_{n_k}(j)\right), \label{eq:graph-pos-enc} \\
    \nonumber \\
    z_k(j) = &f_2( g_k(1,j), g_k(2,j), \ldots, g_k(j,j), \nonumber \\
     &h_k(1,j), h_k(2,j), \ldots, h_k(j,j))
\end{align}
\end{small}
where $f_1$ is a one-layer MLP (a linear projection with a bias) and $f_2$ is a two layers MLP with a ReLU activation function in the first layer and a sigmoid activation function in the second layer. Also, $g_k$ and $h_k$ are given in Equations~\eqref{eq:normalized-num-paths-1} and \eqref{eq:normalized-num-paths-2}.
    
\subsection{Avoiding Isolated Nodes} \label{sec:avoidAllZeros}

 When generating the $k$-th node, our model should decide its connection to the preceding $(k-1)$ nodes. In other words, it should try to estimate the first $(k-1)$ elements of $L_k^{\pi}$ (because the other elements of this vector are equal to zero according to the definition.)
 Since we use BFS orders for training, we should generate graphs in the BFS sequences. However, when we consider a BFS ordering of the nodes of a graph, each node like $v$ (except the root node) has a parent node in the BFS tree ($p_v$). In this ordering, $p_v$ appears at a position before $v$. Therefore, adding an isolated node, which does not connect to any of the preceding nodes, is not allowed.
 However, our model so far has nothing to satisfy this condition. You et al. \cite{graphRNN} allows the isolated nodes to be generated and remove them after generating the graph. However, it will disturb the distribution of the generated graph sizes. In this section, we tackle this problem by distributing the probability of the all-zeros vector among valid vectors. Let the binary vector $\textbf{y} = \langle y_1, y_2, \ldots, y_{k-1} \rangle$ denote the model estimation of the first $k-1$ bits of $L_k^{\pi}$, and $P(\textbf{y})$ denote the probability of generation of $\textbf{y}$. 

We define the new density function $\tilde{P}$ over $k-1$-bits binary vectors as follows:

\begin{small}
\begin{align}\label{eq:P_vector_noAllZeros}
    \tilde{P}(\textbf{y}) = \begin{cases}
    0 & \hspace{-0.8cm}\text{if $y_1 = \ldots = y_{k-1} = 0$}\\
    \frac{P(\textbf{y})}{1 - P(\overline{0})} &\text{otherwise}
    \end{cases}
\end{align}
\end{small}

Computation of the new loss function (the new negative log-likelihood) during the training process is easy; it just needs an extra computation of $P(\overline{0})$. However, generation is a bit challenging. A simple method for generating samples from $\tilde{P}$ is generating samples from $P$ and rejecting all-zeros samples. But this sampling method is likely to be trapped in long-lasting rejection loops, especially for generating sparse graphs.
In order to sample from $\Tilde{P}(\textbf{y})$ sequentially, it is sufficient to compute $\Tilde{P}(Y_i \vert y_1, y_2, \ldots, y_{i-1})$. We prove the following theorem to show how we can efficiently sample binary vectors from $\tilde{P}$.

\begin{theorem} \label{theo:sampling}
For a fixed set of values for $y_1$, $y_2$, \ldots, $y_{i-1}$, let  $P(Y_i =1 \vert y_1, \ldots, y_{i-1})$ denoted by $P_i$ then
\begin{footnotesize}
\begin{align}
 \tilde{P}(Y_i=1 \vert y_1,\ldots y_{i-1})&= 
	 \begin{cases}
       P_i & \text{if~~} \exists j<i  \\
        & \text{s.t.~~} y_j=1 \\
       \frac{P_i \prod\limits_{1 \le j < i}P(y_j \vert y_1,\ldots,y_{j-1})} {\prod\limits_{1 \le j < i}P(y_j \vert y_1,\ldots,y_{j-1}) - P(\overline{0})} & \text{otherwise.}
    \end{cases}
\end{align}
\end{footnotesize}	 
For the special case of $i=1$,
\begin{small}
\begin{align}
    \tilde{P}(Y_1=1) &= P(Y_1=1) \times \frac{1}{1-P(\overline{0})}.
\end{align}
\end{small}
\end{theorem}

\begin{proof}
We first define the following sets:
\begin{small}
\begin{align}
    A_y^{(i)} &\triangleq \lbrace \langle a_1, \ldots, a_{k-1} \rangle \in \lbrace 0, 1 \rbrace^{k-1} \vert \forall j < i: a_j = y_j \rbrace \\
	A_0^{(i)} &\triangleq \lbrace \langle a_1, \ldots, a_{k-1} \rangle \in \lbrace 0, 1 \rbrace^{k-1} \vert  a_i = 0 \rbrace \\ 
	A_1^{(i)} &\triangleq \lbrace \langle a_1, \ldots, a_{k-1} \rangle \in \lbrace 0, 1 \rbrace^{k-1} \vert  a_i = 1 \rbrace 
\end{align}
\end{small}
In other words, $A_y^{(i)}$ is the set of all binary sequences of length $k-1$ whose first $i-1$ elements are equal to $y_1, y_2, \ldots, y_{i=1}$. $A_0^{(i)}$ represents all binary sequences of length $k-1$ whose $i$-th element is equal to $0$, and $A_1^{(i)}$ represents all binary sequences of length $k-1$ whose $i$-th element is equal to $1$.
As a result, it follows that

\begin{small}
\begin{align} 
    \tilde{P}(Y_i=1 \vert y_1, \ldots y_{i-1}) &= \frac{\tilde{P}(A_y^{(i)} \cap A_1^{(i)})}{\tilde{P}(A_y^{(i)})} \label{eq:AcA0A1fraction}\\
    &=\frac{\tilde{P}(A_y^{(i)} \cap A_1^{(i)})}{\tilde{P}(A_y^{(i)} \cap A_0^{(i)}) + \tilde{P}(A_y^{(i)} \cap A_1^{(i)})}\nonumber. 
\end{align}
\end{small}
Since $A_y^{(i)} \cap A_1^{(i)}$ does not include the zero vector, and respecting the definition of the probability mass function $\tilde{P}(.)$ in Equation \eqref{eq:P_vector_noAllZeros}, we infer that

\begin{small}
\begin{align}
    \tilde{P}(A_y^{(i)} \cap A_1^{(i)}) &= \displaystyle \sum_{c \in A_y^{(i)} \cap A_1^{(i)}} \tilde{P}(c) = \sum_{c \in A_y^{(i)} \cap A_1^{(i)}} \frac{P(c)}{1-P(\overline{0})} \\ 
    &= \frac{P(A_y^{(i)} \cap A_1^{(i)})}{1-P(\overline{0})} \nonumber \\
    &=\frac{1}{1-P(\overline{0})}P(Y_i=1\vert y_1,\ldots,y_{i-1})  \nonumber \\ 
    & \times \prod\limits_{1 \le j < i}P(y_j \vert y_1,\ldots,y_{j-1}). \label{eq:A1}
\end{align}
\end{small}
If at least one of $y_1, \ldots, y_{i-1}$ is equal to $1$, then $A_y^{(i)} \cap A_0^{(i)}$ does not include the zero vector as well, and we can infer in a similar way that

\begin{small}
\begin{align}
    \tilde{P}(A_y^{(i)} \cap A_0^{(i)})
    &= \frac{1}{1-P(\overline{0})}P(Y_i=0\vert y_1,\ldots,y_{i-1}) \nonumber \\ 
    & \times \prod\limits_{1 \le j < i}P(a_j \vert y_1,\ldots,y_{j-1}).
\end{align}
\end{small}
Hence, respecting Equation \eqref{eq:AcA0A1fraction}, we conclude that if there exists $j < i$  such that for $y_j=1$, then we have 

\begin{small}
\begin{align} 
    \tilde{P}(Y_i=1 \vert y_1, \ldots y_{i-1}) &= 
    \hspace{0.2cm}P(Y_i=1 \vert y_1, \ldots y_{i-1}),  \label{eq:Ptilde-1}
\end{align}
\end{small}
which equals to $P_i$ by definition.
But if $A_y^{(i)} \cap A_0^{(i)}$ contains the zero vector, it means that
$y_1=y_2=\ldots=y_{i-1}=0.$  In this case, we have:
\begin{small}
\begin{align}
    \tilde{P}(A_y^{(i)} \cap A_0^{(i)}) &= \sum_{c \in A_y^{(i)} \cap A_0^{(i)}} \tilde{P}(c) = \sum_{\substack{c \in A_y^{(i)} \cap A_0^{(i)},\\ c \ne \overline{0}}} \frac{P(c)}{1-P(\overline{0})} \nonumber \\
    &= \frac{P(A_y^{(i)} \cap A_0^{(i)}) - P(\overline{0})}{1-P(\overline{0})}\nonumber \\ 
    &= \prod\limits_{1 \le j < i}P(y_j \vert y_1,\ldots,y_{j-1}) \nonumber \\
    & \times \frac{P(Y_i=0\vert y_1,\ldots,y_{i-1})}{1-P(\overline{0})} - \frac{P(\overline{0})}{1-P(\overline{0})}. \label{eq:A0}
\end{align}
 \end{small}
Putting Equations \eqref{eq:A1} and \eqref{eq:A0} into Equation \eqref{eq:AcA0A1fraction}, it follows that when  $y_1=\ldots=y_{i-1}=0$, then we have 

\begin{small}
\begin{align} 
    \tilde{P}(Y_i=1 \vert y_1, \ldots y_{i-1}) &=
     \frac{P_i\prod\limits_{1 \le j < i}P(y_j \vert y_1,\ldots,y_{j-1})}{\prod\limits_{1 \le j < i}P(y_j \vert y_1,\ldots,y_{j-1}) - P(\overline{0})} \label{eq:Ptilde-2}
\end{align}
\end{small}
For the special case of $i=1$, it is clear that
$\tilde{P}(Y_1 = 1) = \tilde{P}(A_1^{(i)})$,
and since $A_1^{(i)}$ does not include the zero vector, according to the definition of the probability mass function $\tilde{P}(.)$ in Equation \eqref{eq:P_vector_noAllZeros} we have
    
\begin{small}
\begin{align}
    \tilde{P}(A_1^{(i)}) &= \sum_{c \in A_1^{(i)}} \tilde{P}(c) \nonumber  
    = \sum_{c \in A_1^{(i)}} P(c) \times \frac{1}{1 - P(\bar{0})} \nonumber \\ 
    &= P(Y_1=1) \times \frac{1}{1 - P(\bar{0})}.
\end{align}
\end{small}
This completes the proof of the theorem.
\end{proof}
    
Therefore, it is obvious that using theorem \ref{theo:sampling} the model can sequentially sample $k-1$ bits $y_1, \ldots, y_{k-1}$ in time of $\mathcal{O}(k)$.
  
\section{Experimental results} \label{sec:experiments}
 
\newcommand{\scl}{0.18}

\begin{table*}[htbp]
\caption{MMD scores of different generative models. In each column, the best result is in bold digits, and the best two results are shaded.}
\label{table:MMD}
\begin{center}
\begin{tabular}{ l  l  l  l  l  l  l l l l}
  \multicolumn{1}{c}{} & \multicolumn{3}{c}{Ego dataset} & \multicolumn{3}{c}{Protein dataset} & 
  \multicolumn{3}{c}{Lobster} \\
  \cmidrule(l{15pt}r{15pt}){2-4} \cmidrule(l{15pt}r{15pt}){5-7}
  \cmidrule(l{15pt}r{15pt}){8-10}
  Method & Deg. & Clus. & Orbit & Deg. & Clus. & Orbit & Deg. & Clus. & Orbit\\
  \midrule
  GraphRNN-S & 0.16 (0.08) & 0.06 (0.06) & 0.13 (0.09) & 0.21 (0.02) & \cellcolor{blue!15}\textbf{0.19 (0.05)} & \cellcolor{blue!15}0.15 (0.02) & \cellcolor{blue!15}0.08 (0.02) & 0.17 (0.05) & \cellcolor{blue!15}0.22 (0.18)\\
  GraphRNN & \cellcolor{blue!15}0.13 (0.03) & 0.50 (0.03) & \cellcolor{blue!15}\textbf{0.03 (0.001)} & \cellcolor{blue!15}0.03 (0.004) & 1.05 (0.01) & 0.53 (0.02) & \cellcolor{blue!15}\textbf{0.01 (0.01)} & \cellcolor{blue!15}\textbf{0.00 (0.00)} & \cellcolor{blue!15}\textbf{0.01 (0.01)}\\
  GRAN & 0.29 (0.03) & \cellcolor{blue!15}\textbf{0.05 (0.02)} & 0.06 (0.01) & 0.04 (0.03) & 0\cellcolor{blue!15}.31 (0.17) & 0.25 (0.07) & 0.56 (0.06) & 0.06 (0.03) & 0.93 (0.14) \\
  Gransformer & \cellcolor{blue!15}\textbf{0.04 (0.001)} & \cellcolor{blue!15}0.08 (0.01) & \cellcolor{blue!15}0.04 (0.002) & \cellcolor{blue!15}\textbf{0.03 (0.003)} & 0.35 (0.03) & \cellcolor{blue!15}\textbf{0.08 (0.01)} & 0.12 (0.07) & \cellcolor{blue!15}\textbf{0.00 (0.00)} & 0.34 (0.22) \\
  \bottomrule
\end{tabular}
\end{center}
\end{table*}

\begin{figure*}[htbp]
    \centering
    \begin{tabular}{cccccc}
         & & & & & \\
         & & & & & \\
         & & & & & \\
         & Original & Gransformer & GRAN & GraphRNN & GraphRNN-S \\
         \rotatebox{90}{Ego}
         & \includegraphics[scale=\scl]{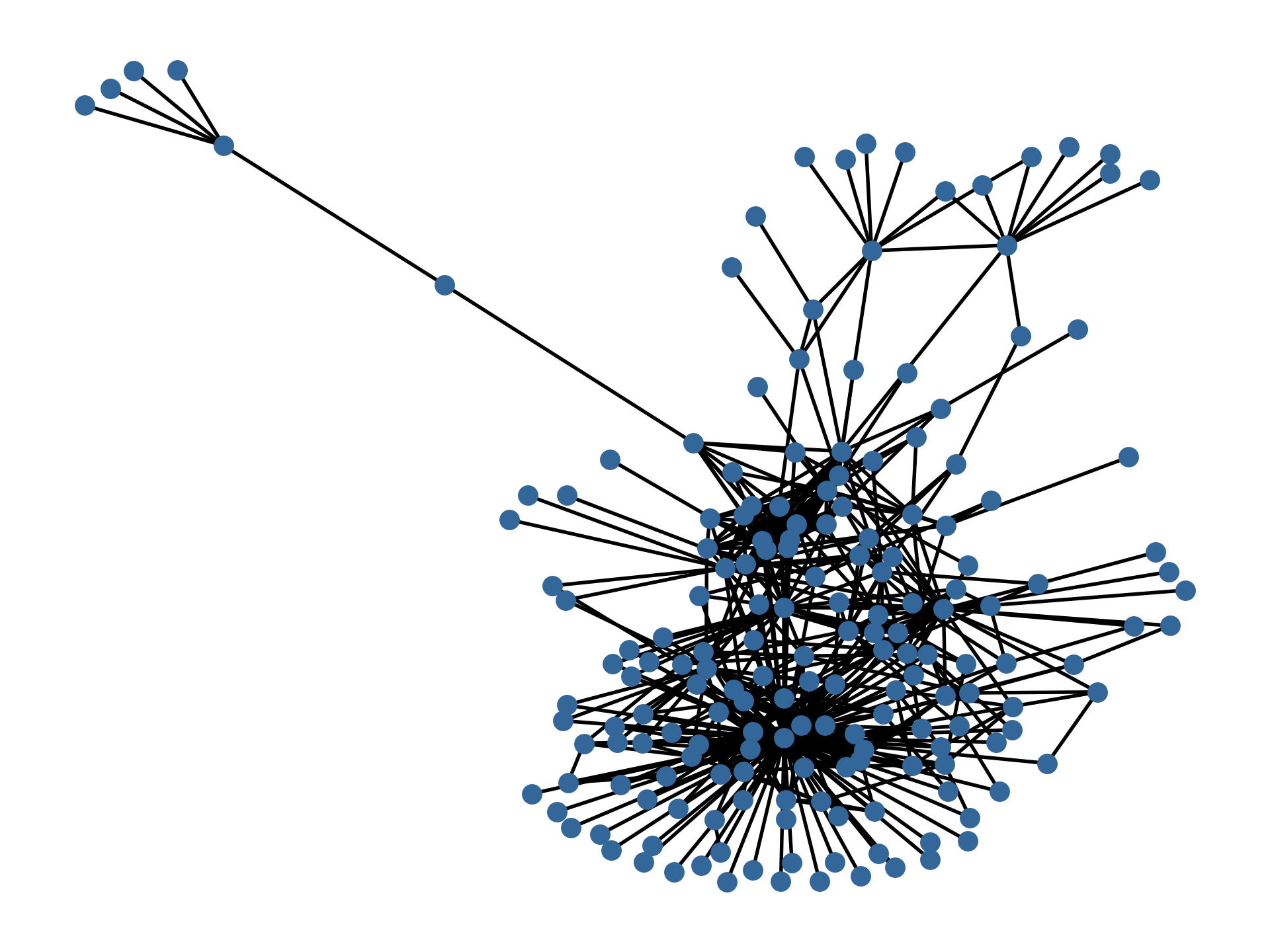} 
         & \includegraphics[scale=\scl]{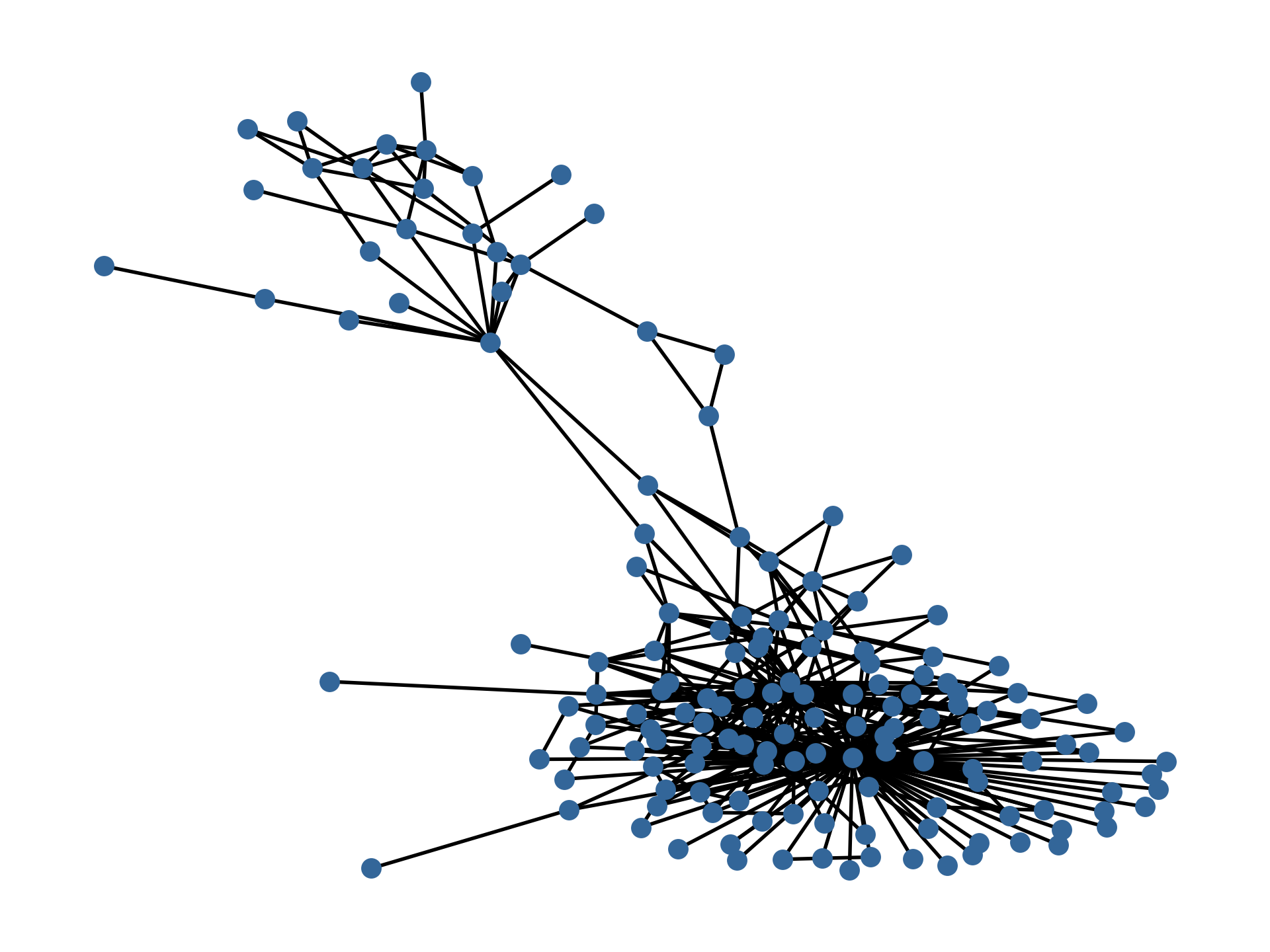} 
         & \includegraphics[scale=\scl]{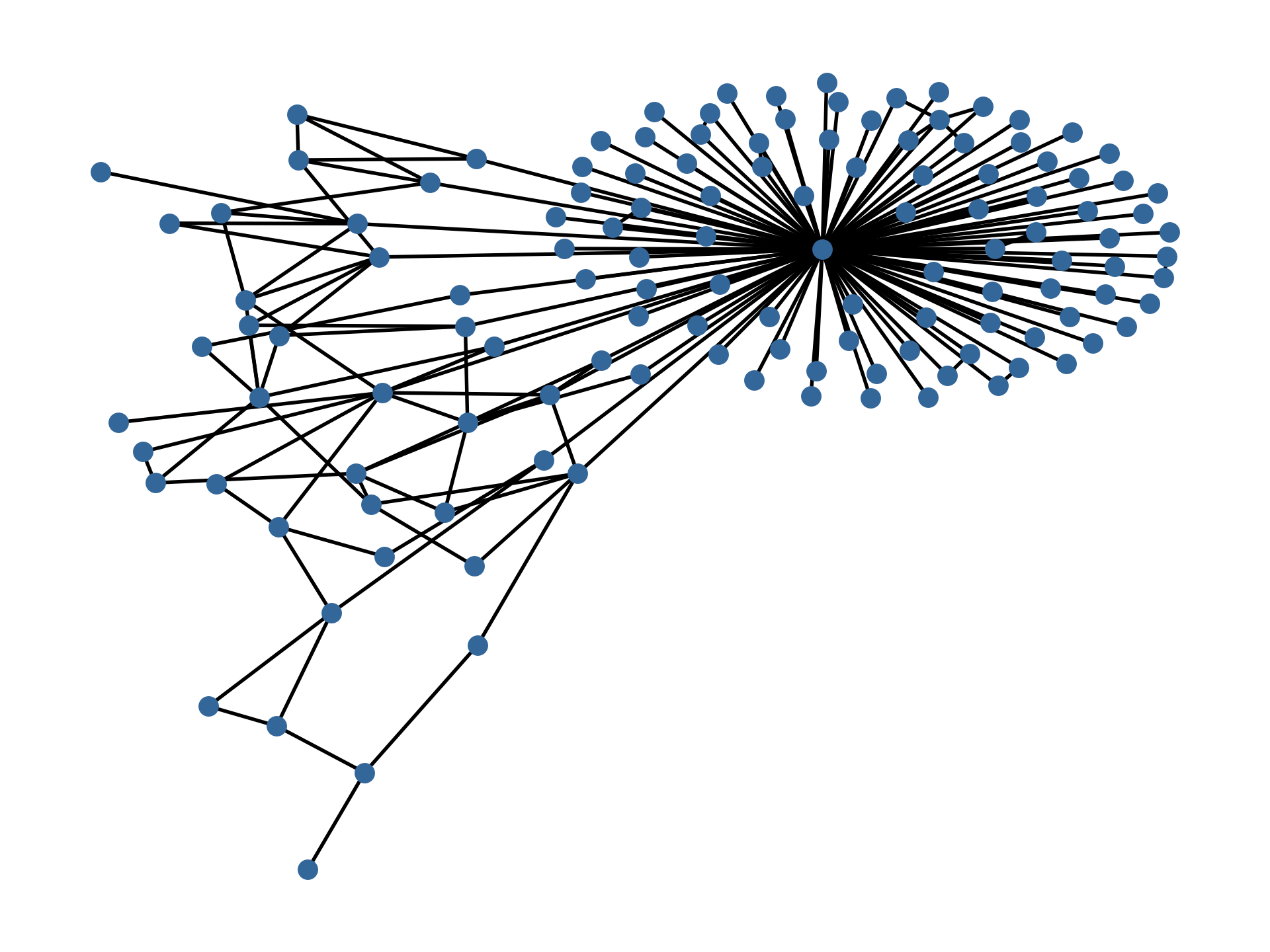}
         &
         \includegraphics[scale=\scl]{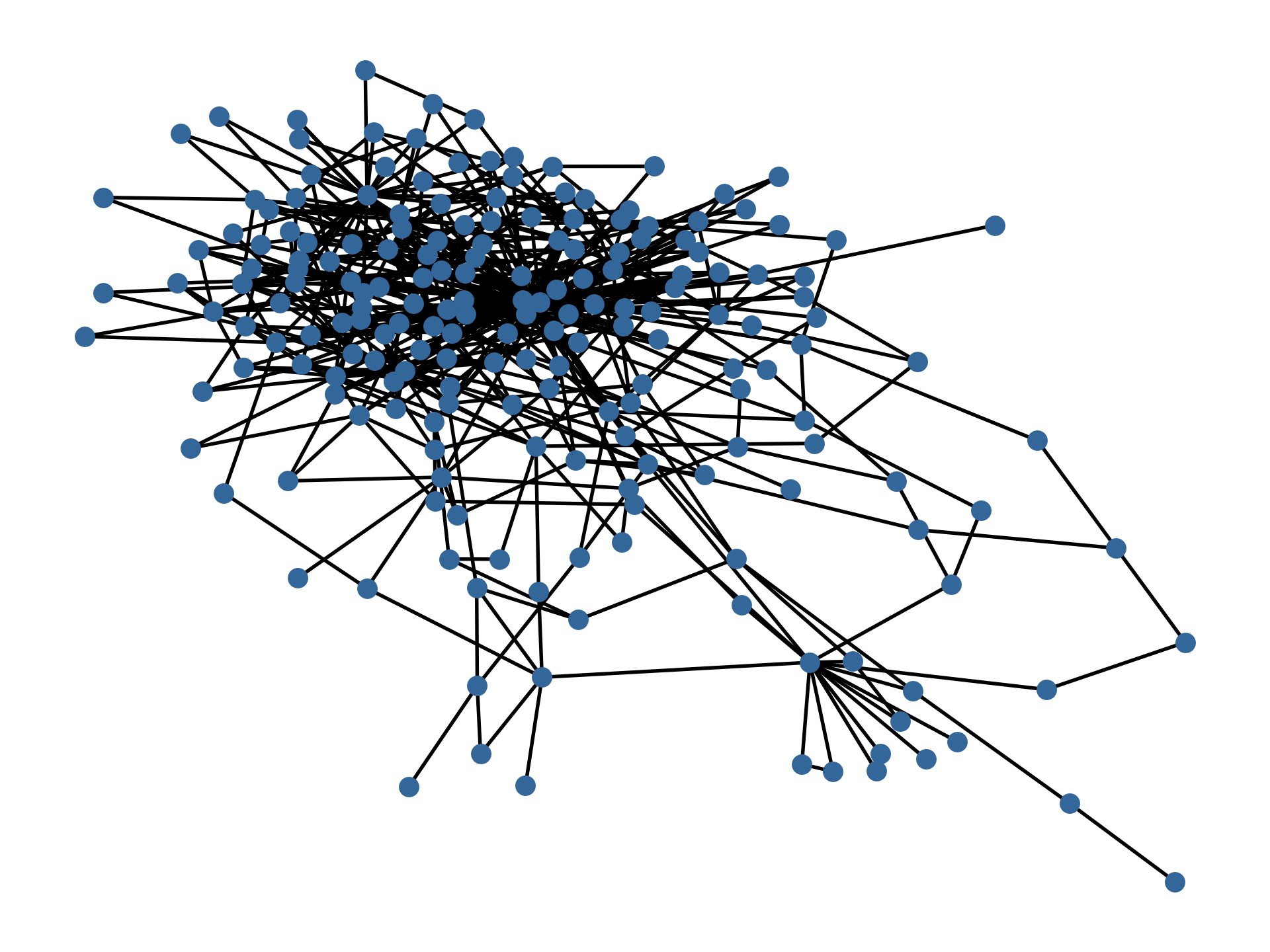}
         &
         \includegraphics[scale=\scl]{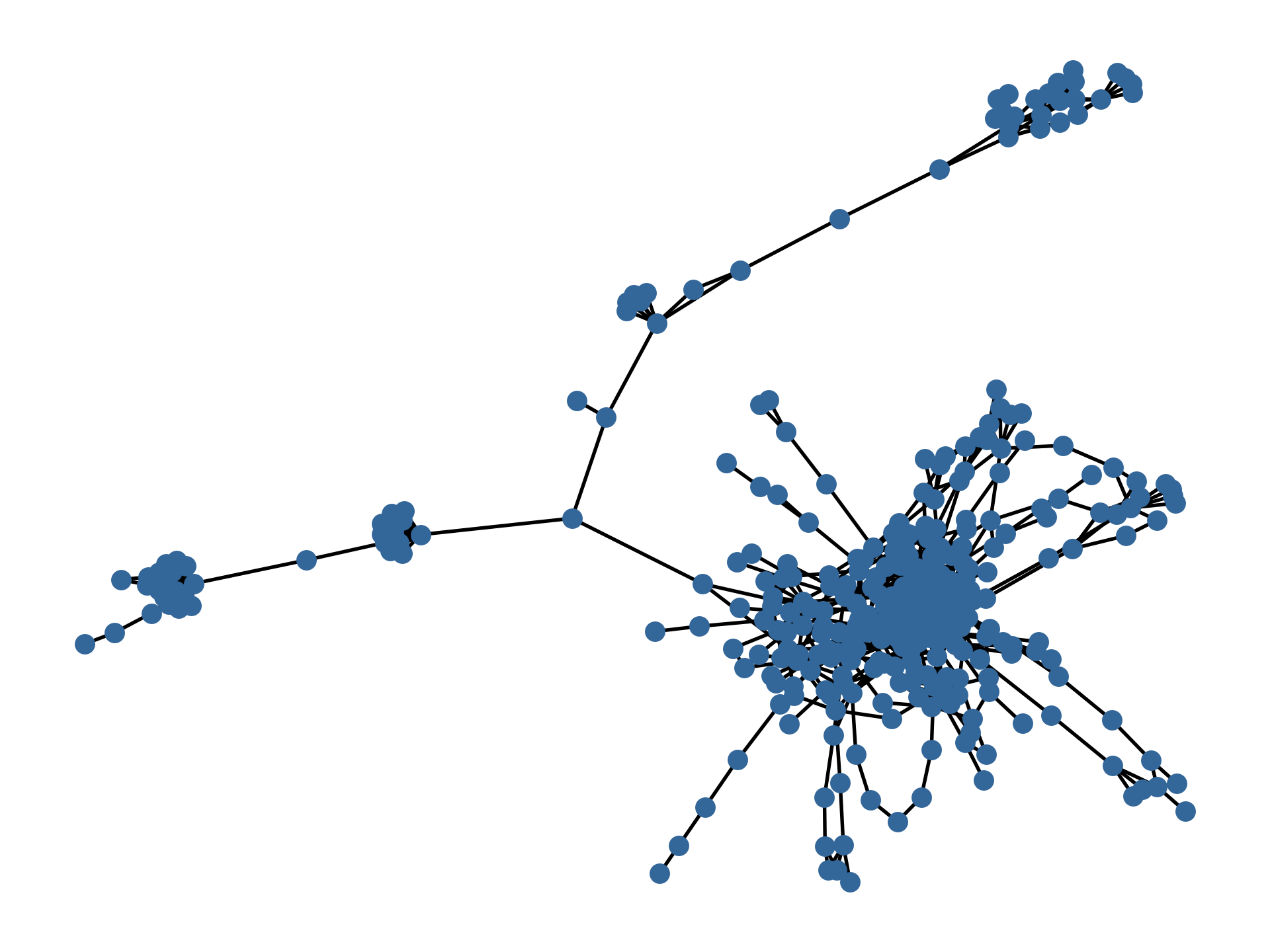}
         \\
         \rotatebox{90}{Ego}
         & \includegraphics[scale=\scl]{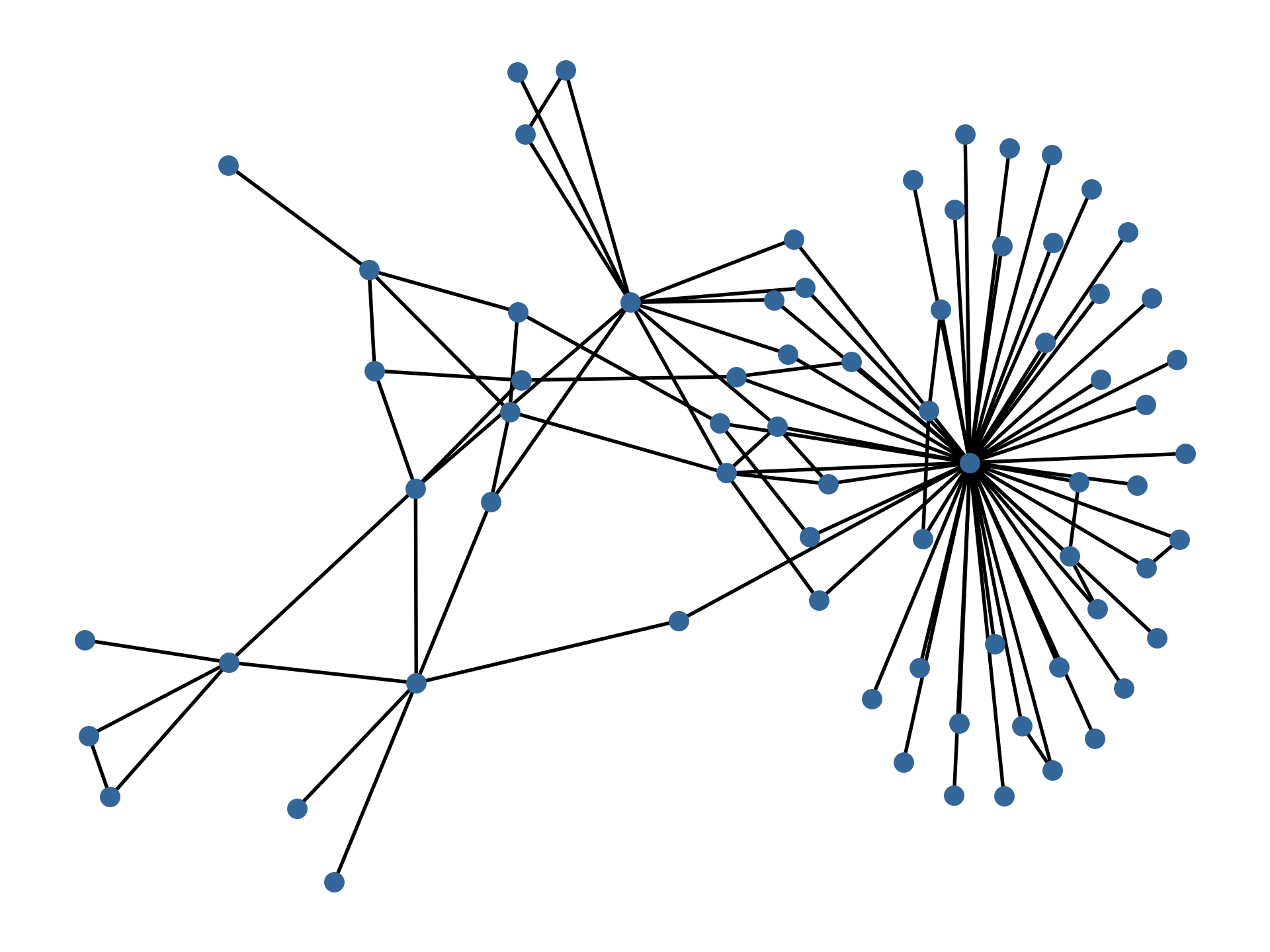}
         & \includegraphics[scale=\scl]{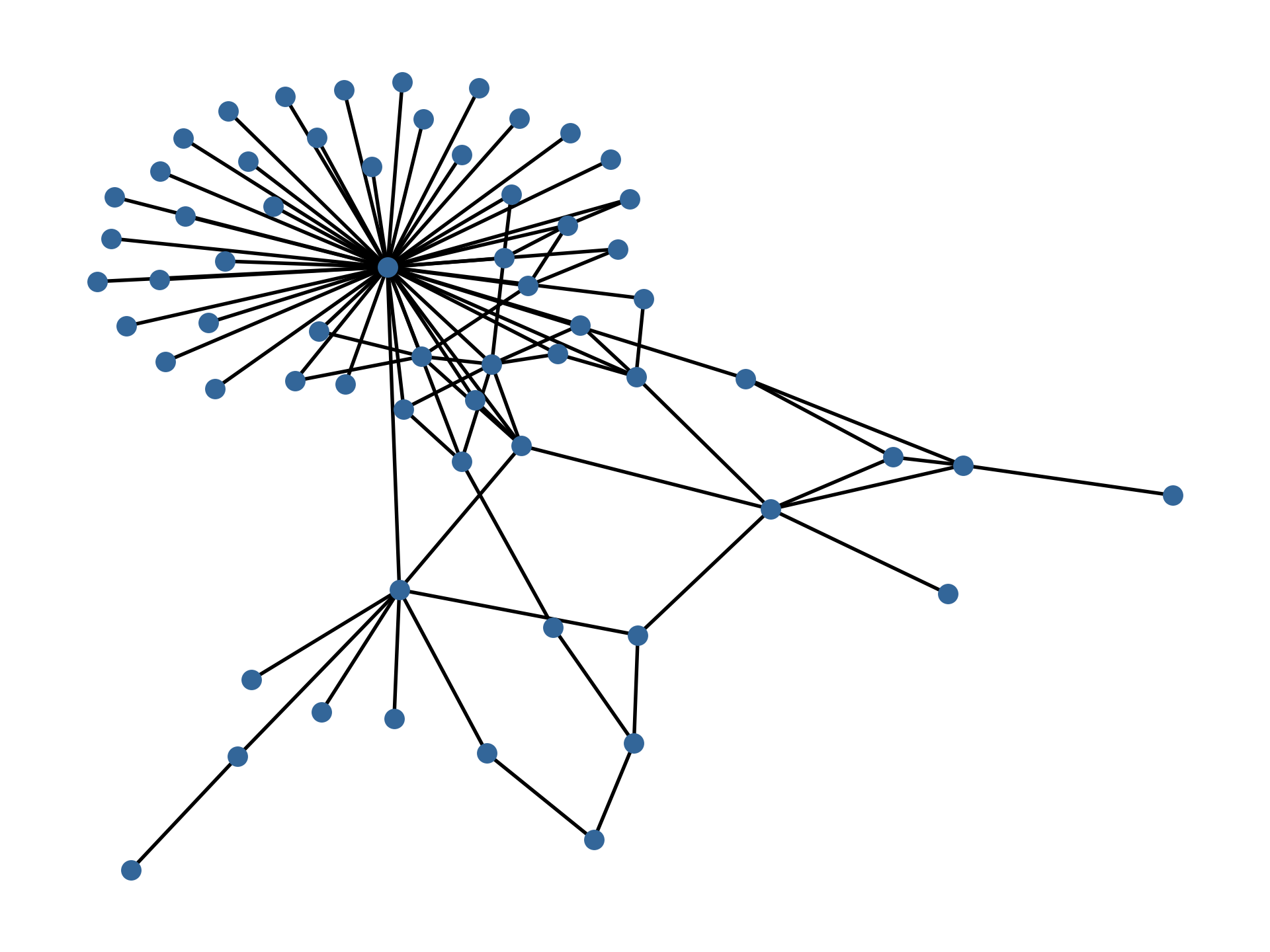} 
         & \includegraphics[scale=\scl]{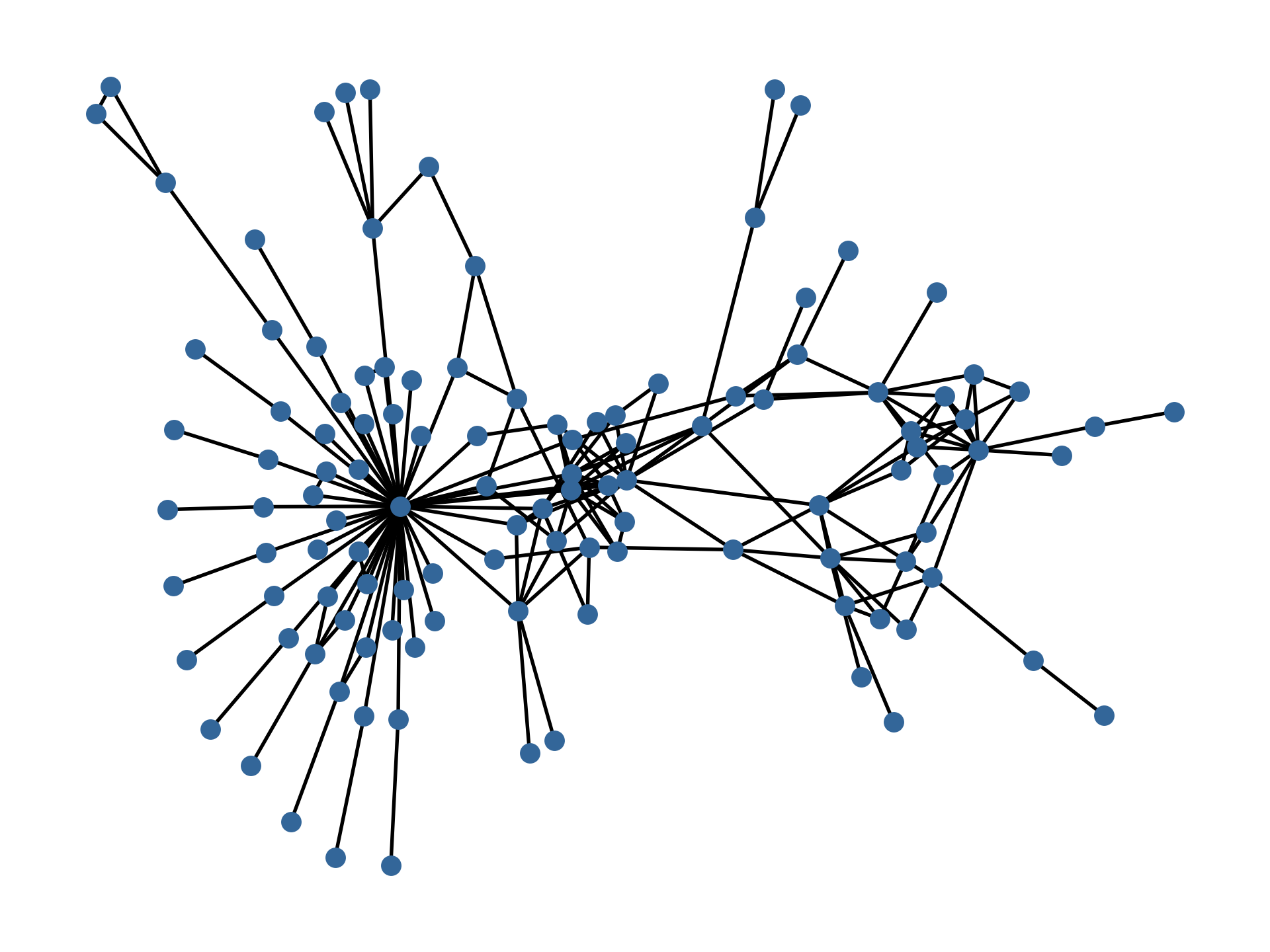}
         &
         \includegraphics[scale=\scl]{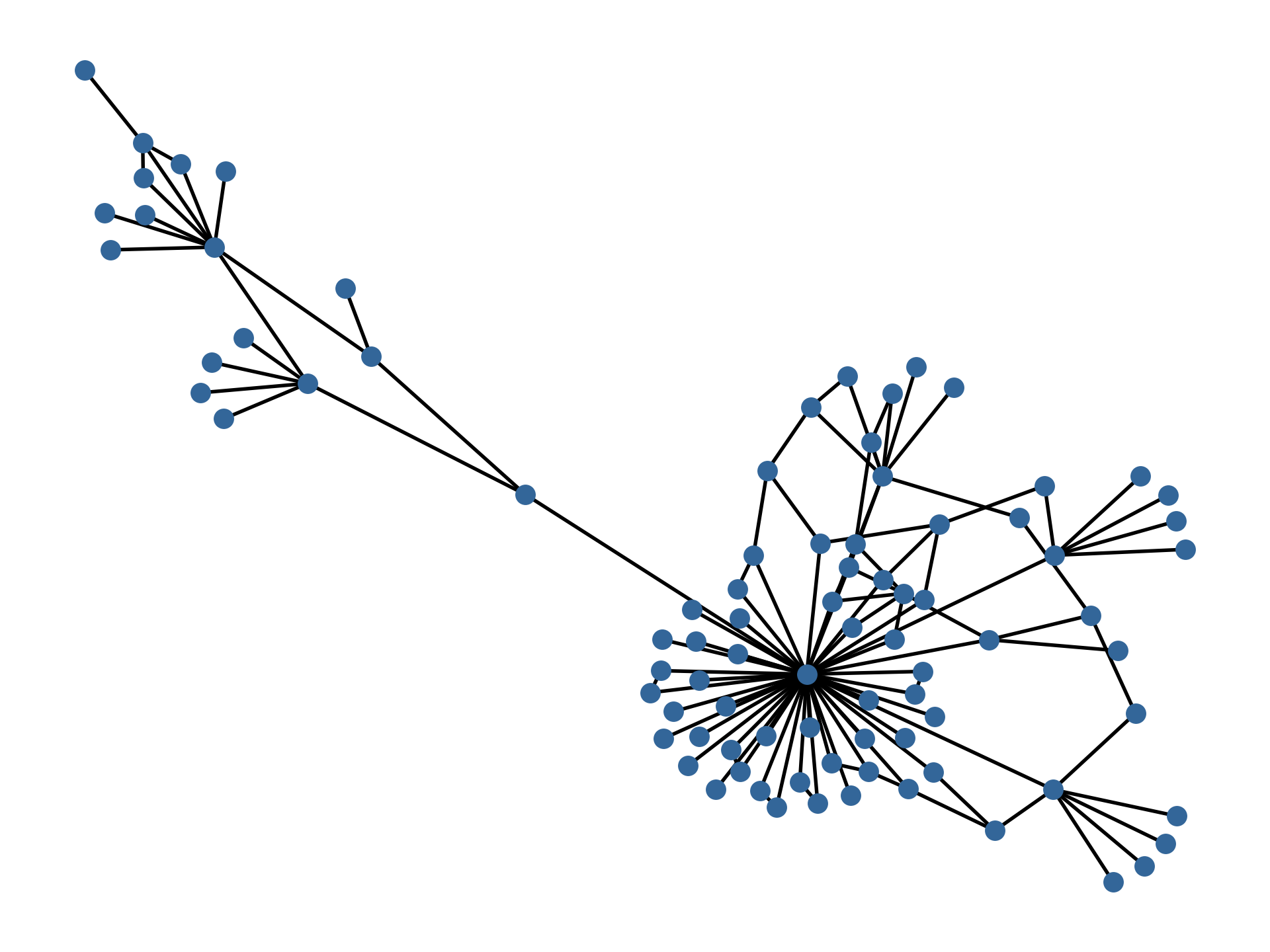}
         &
         \includegraphics[scale=\scl]{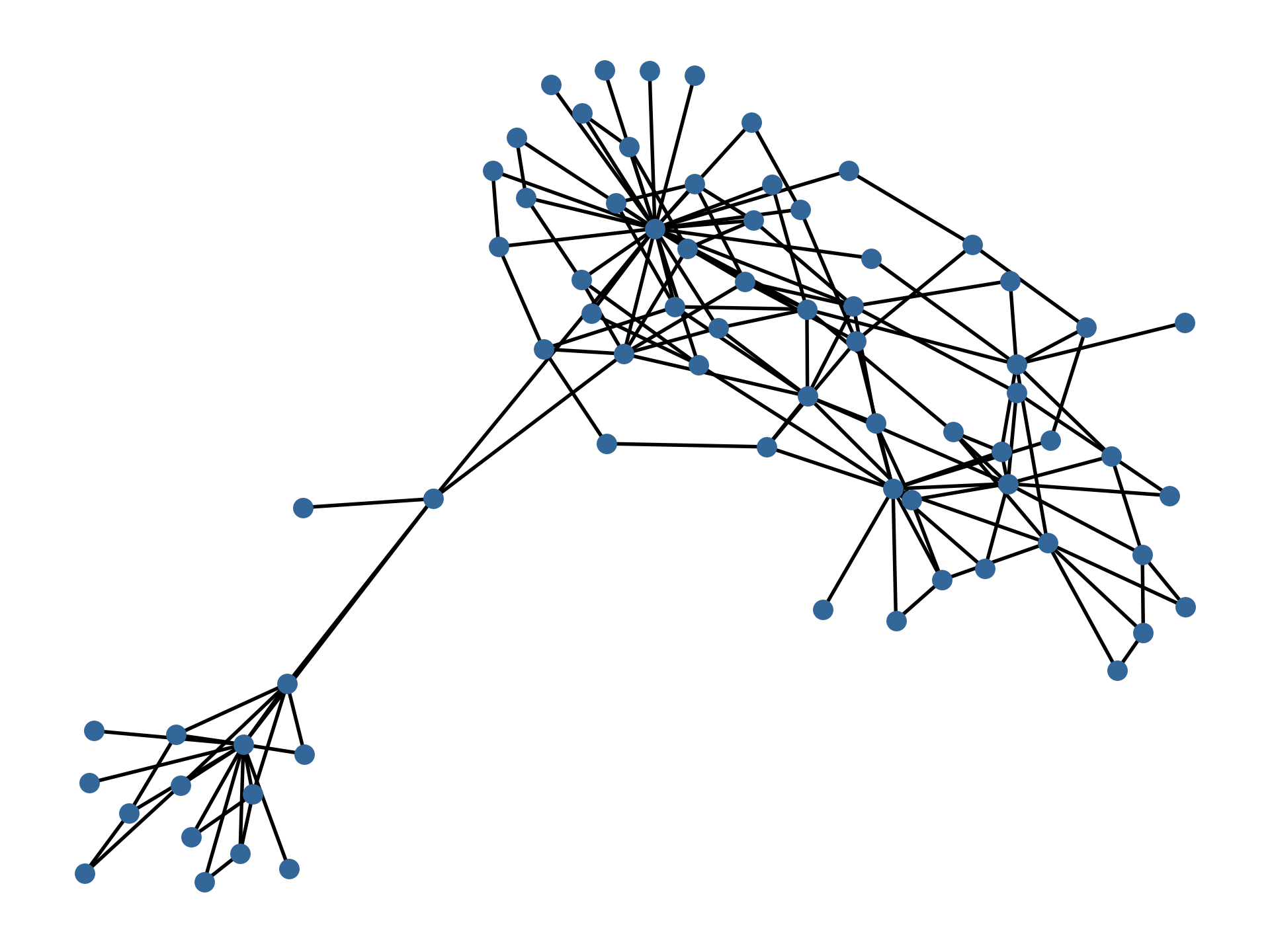}
         \\
         \rotatebox{90}{Protein}
         & \includegraphics[scale=\scl]{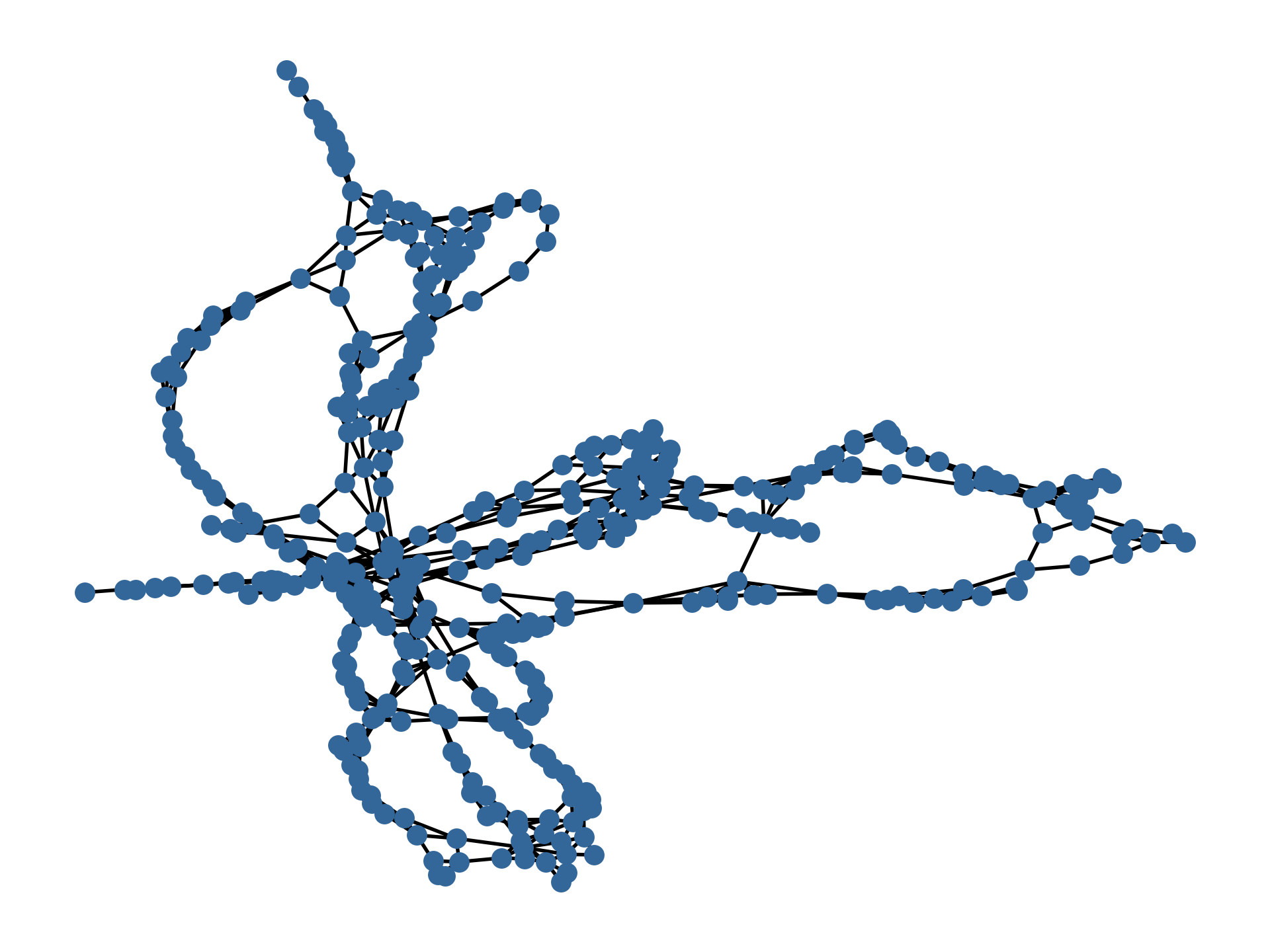} 
         &  \includegraphics[scale=\scl]{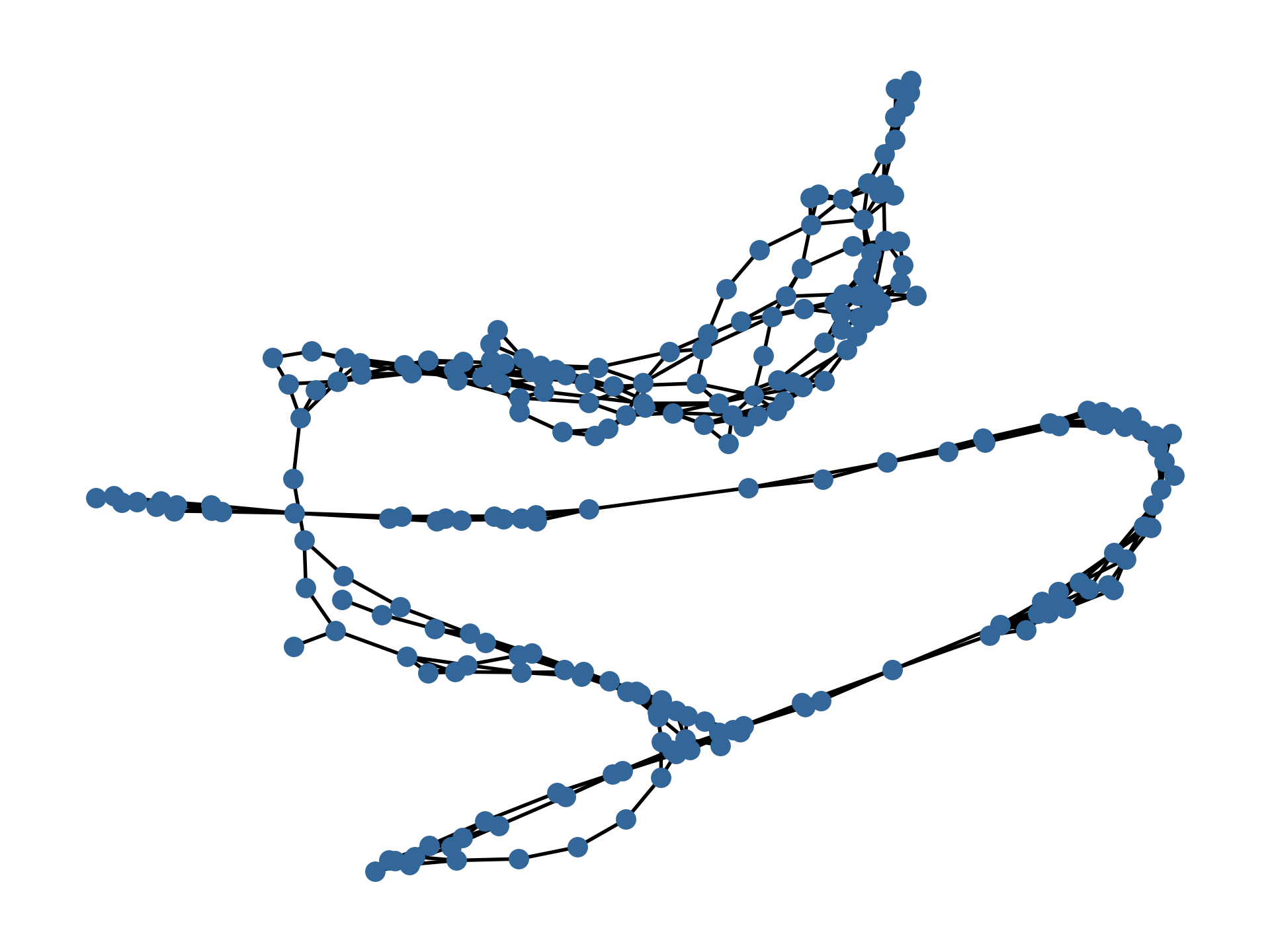} 
         & \includegraphics[scale=\scl]{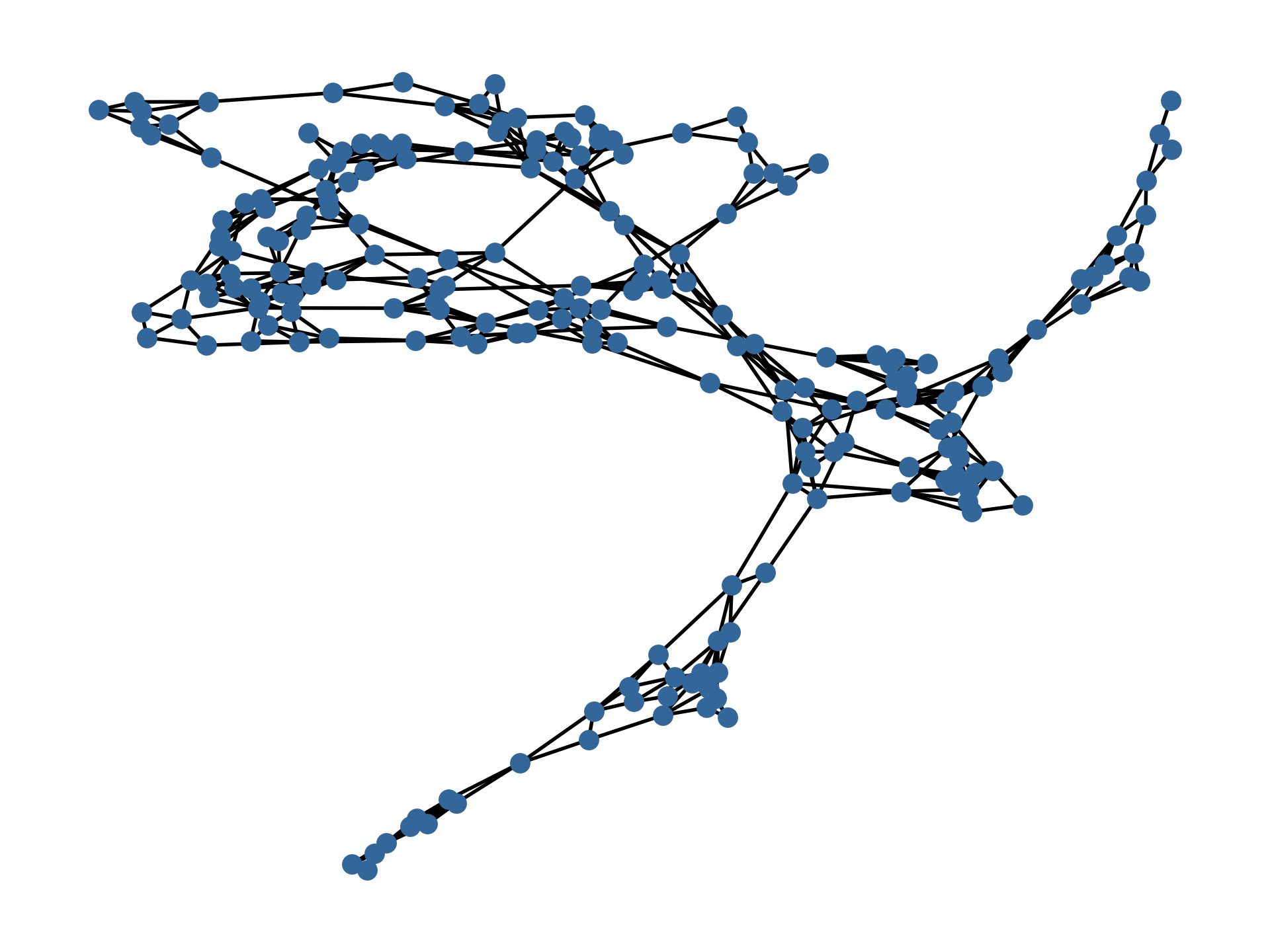}
         &
         \includegraphics[scale=\scl]{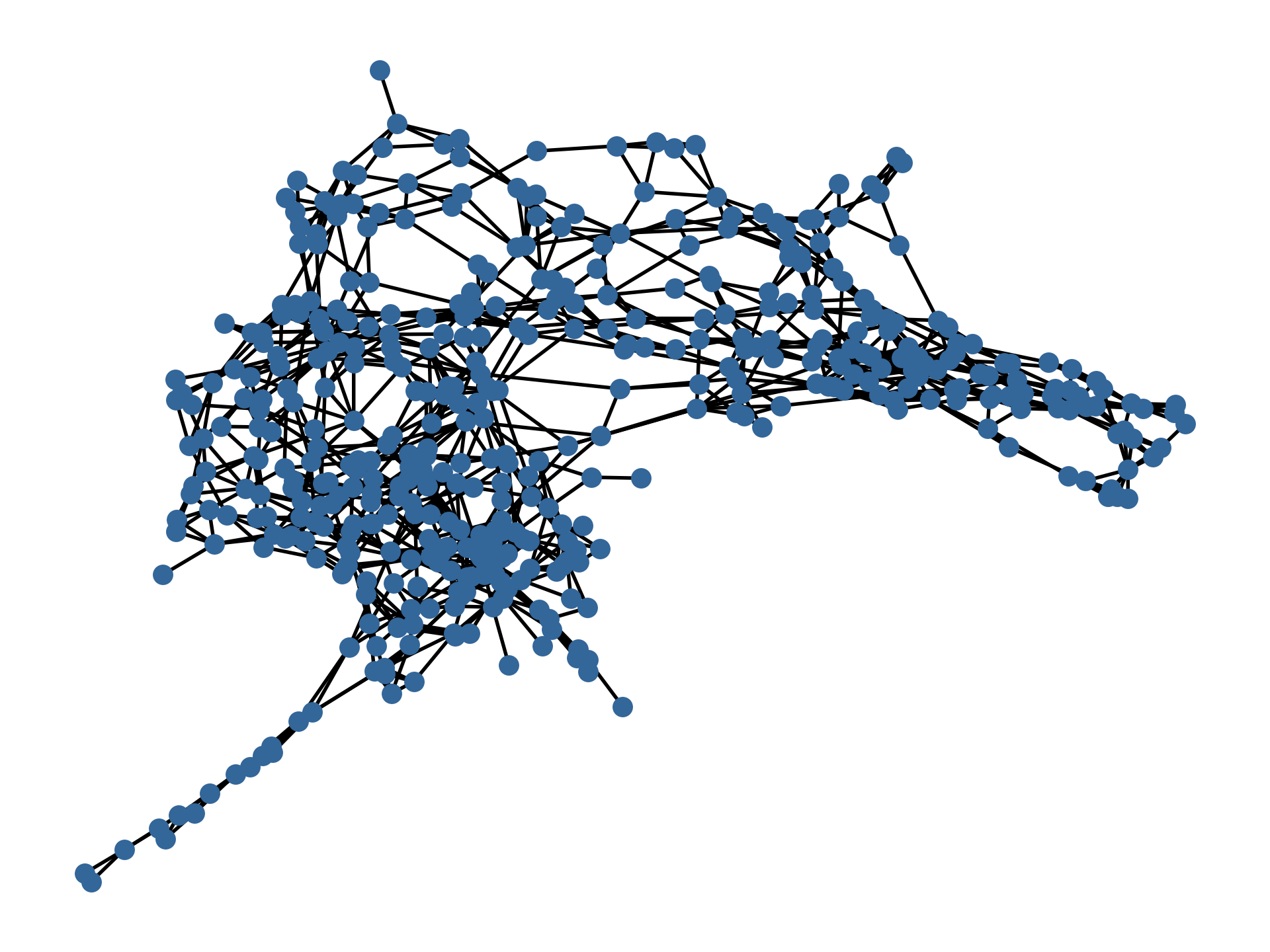}
         &
         \includegraphics[scale=\scl]{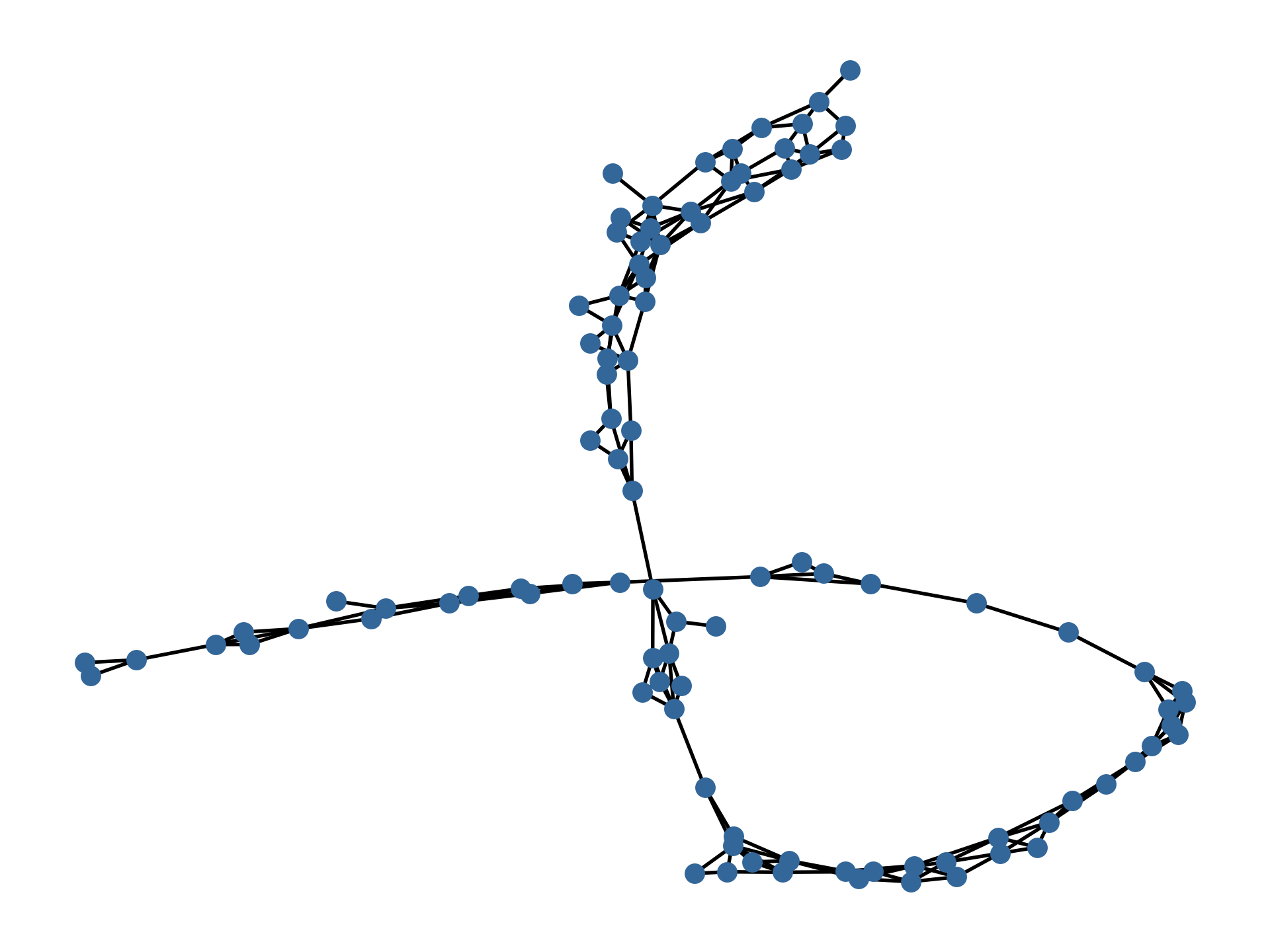}
         \\
         \rotatebox{90}{Protein}
         & \includegraphics[scale=\scl]{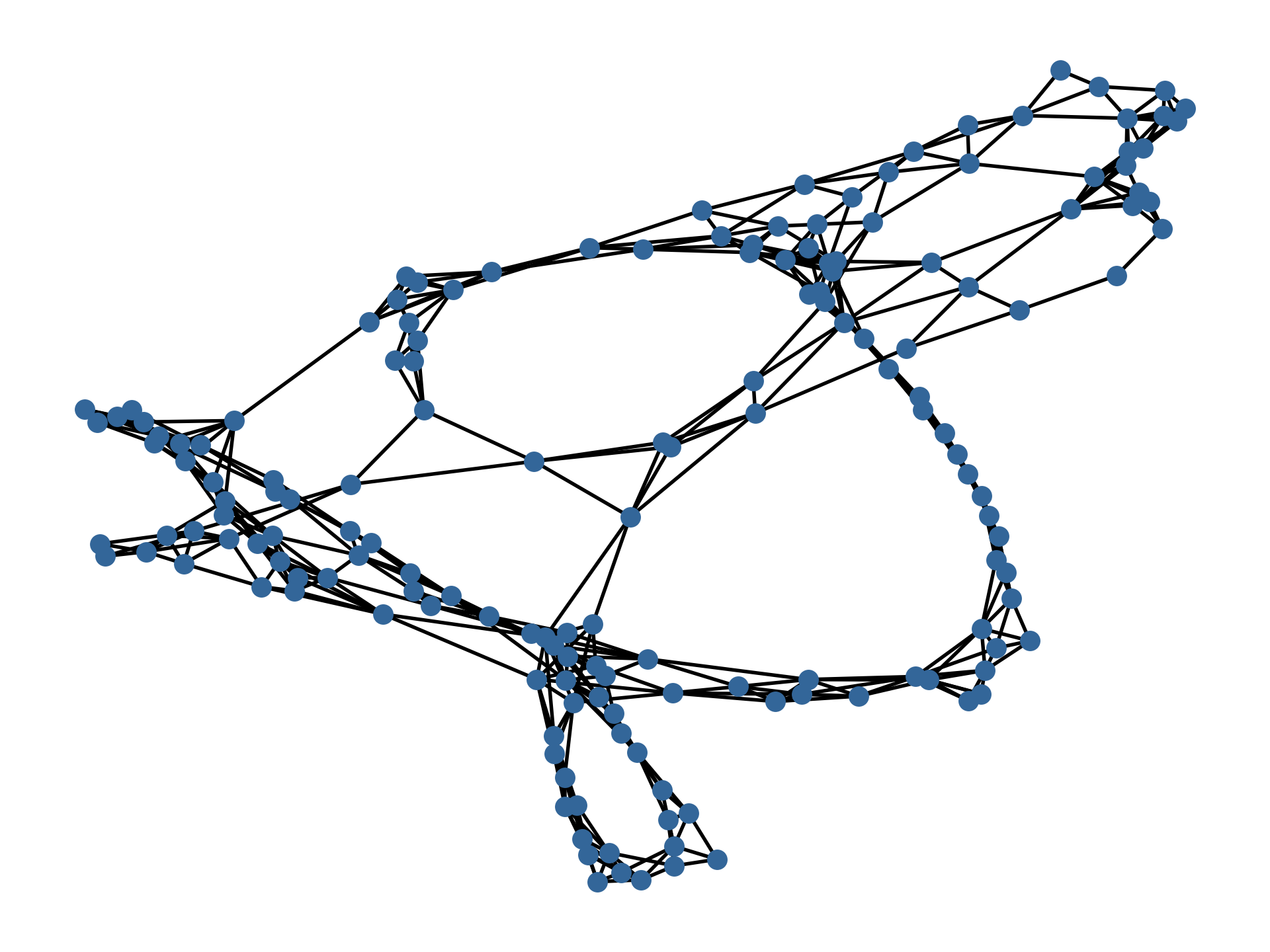} 
         &  \includegraphics[scale=\scl]{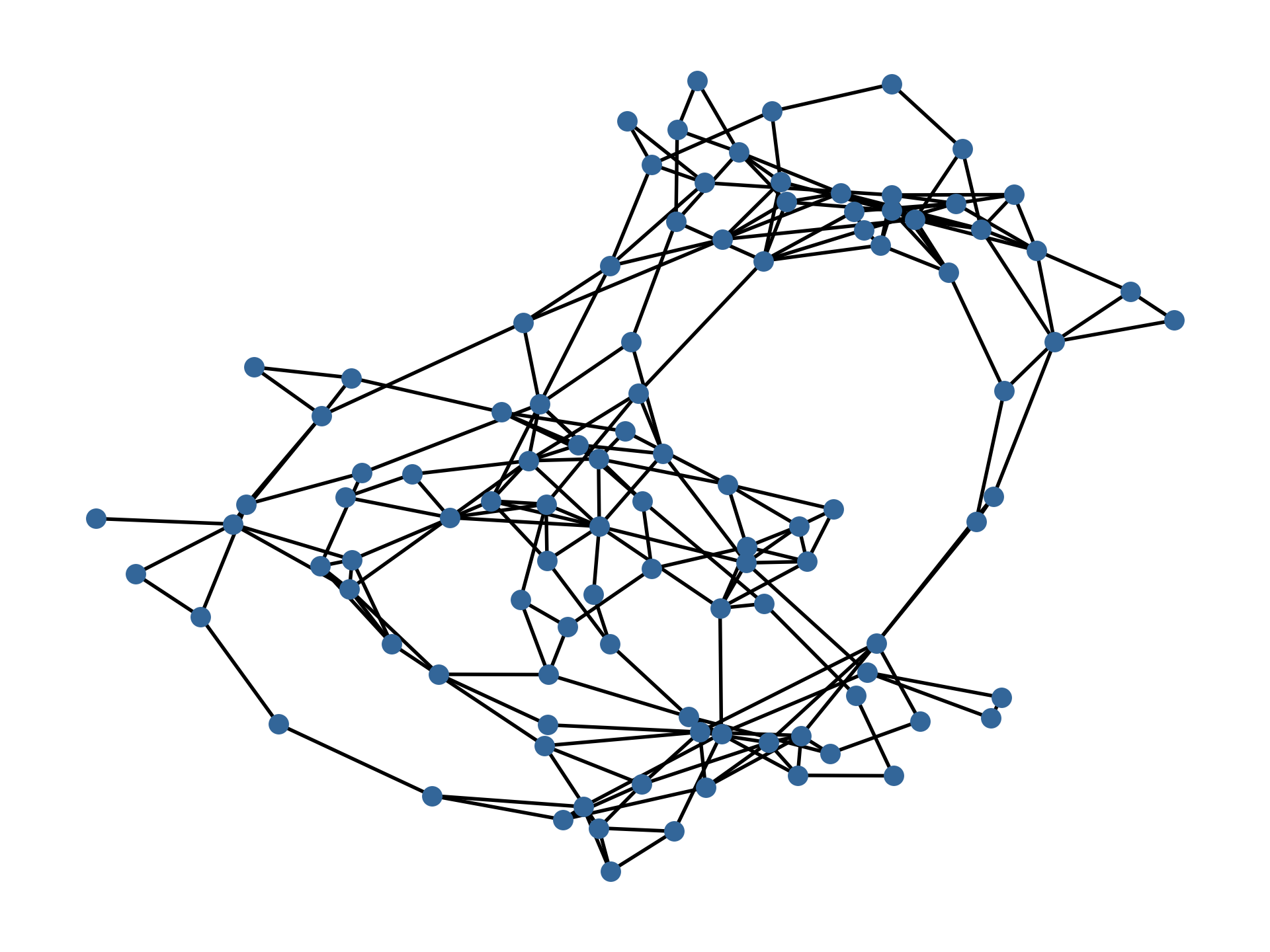} 
         & \includegraphics[scale=\scl]{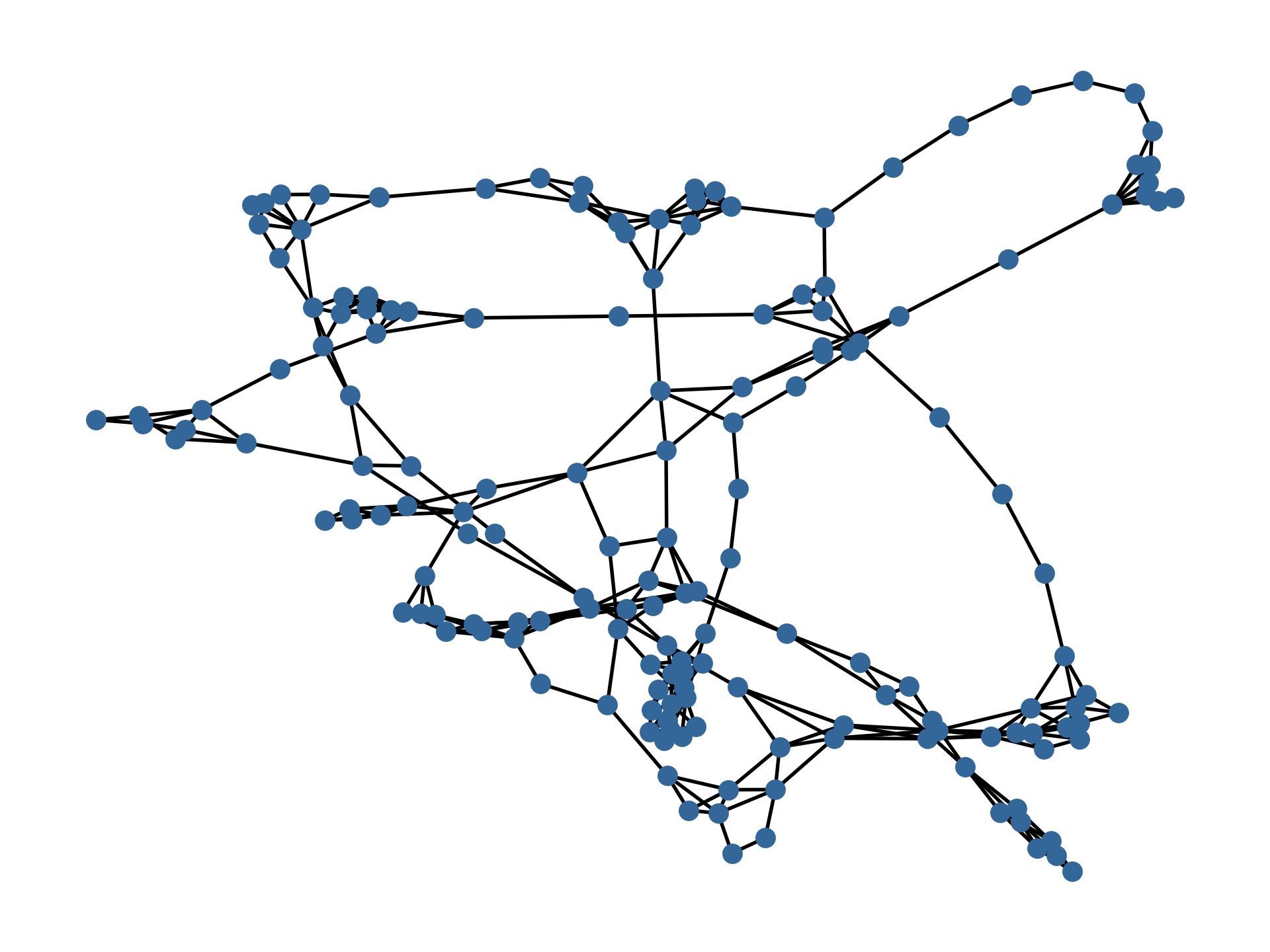}
         &
         \includegraphics[scale=\scl]{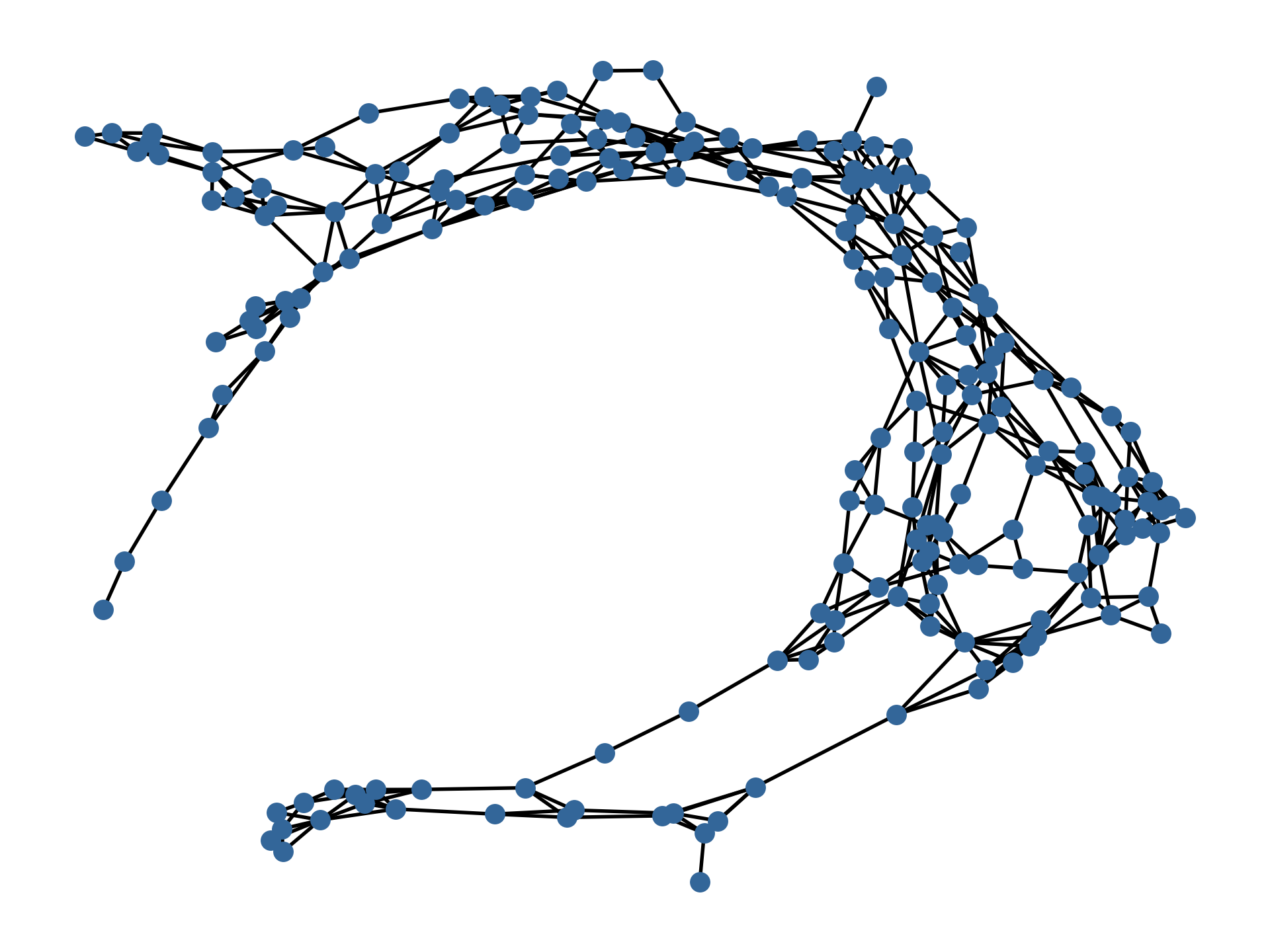}
         &
         \includegraphics[scale=\scl]{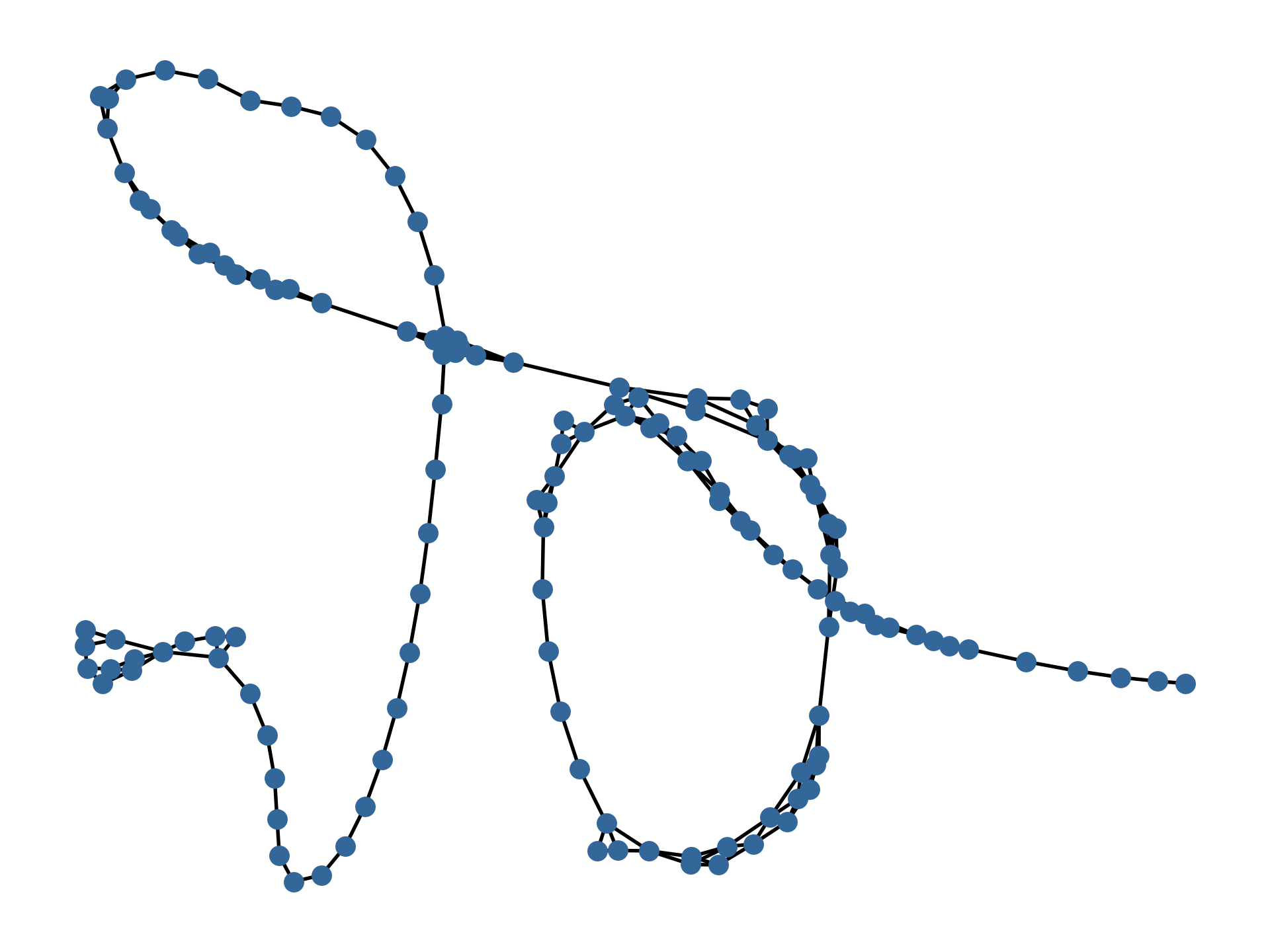}
         \\
         \rotatebox{90}{Lobster}
         & \includegraphics[scale=\scl]{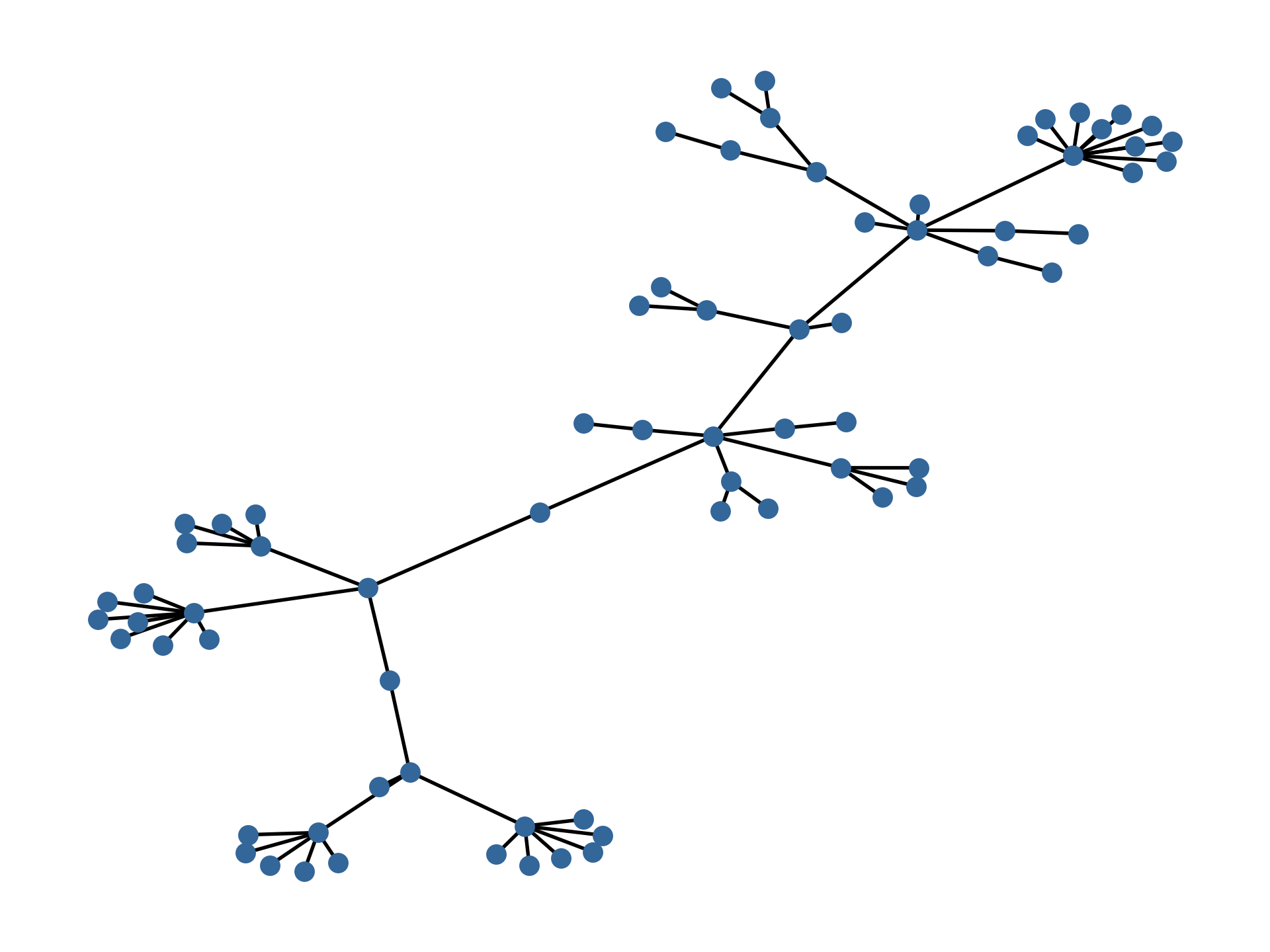} 
         &  \includegraphics[scale=\scl]{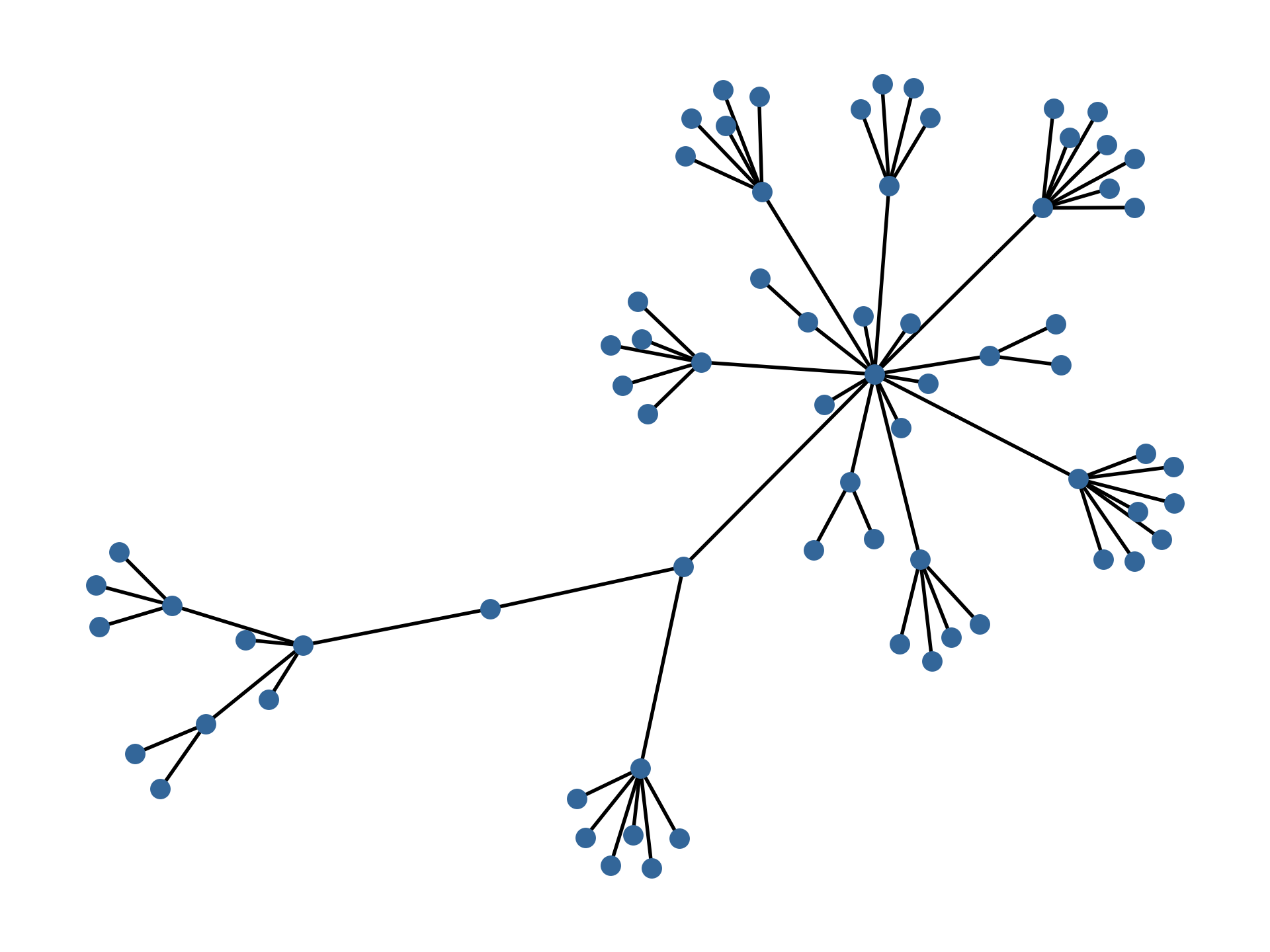} 
         & \includegraphics[scale=\scl]{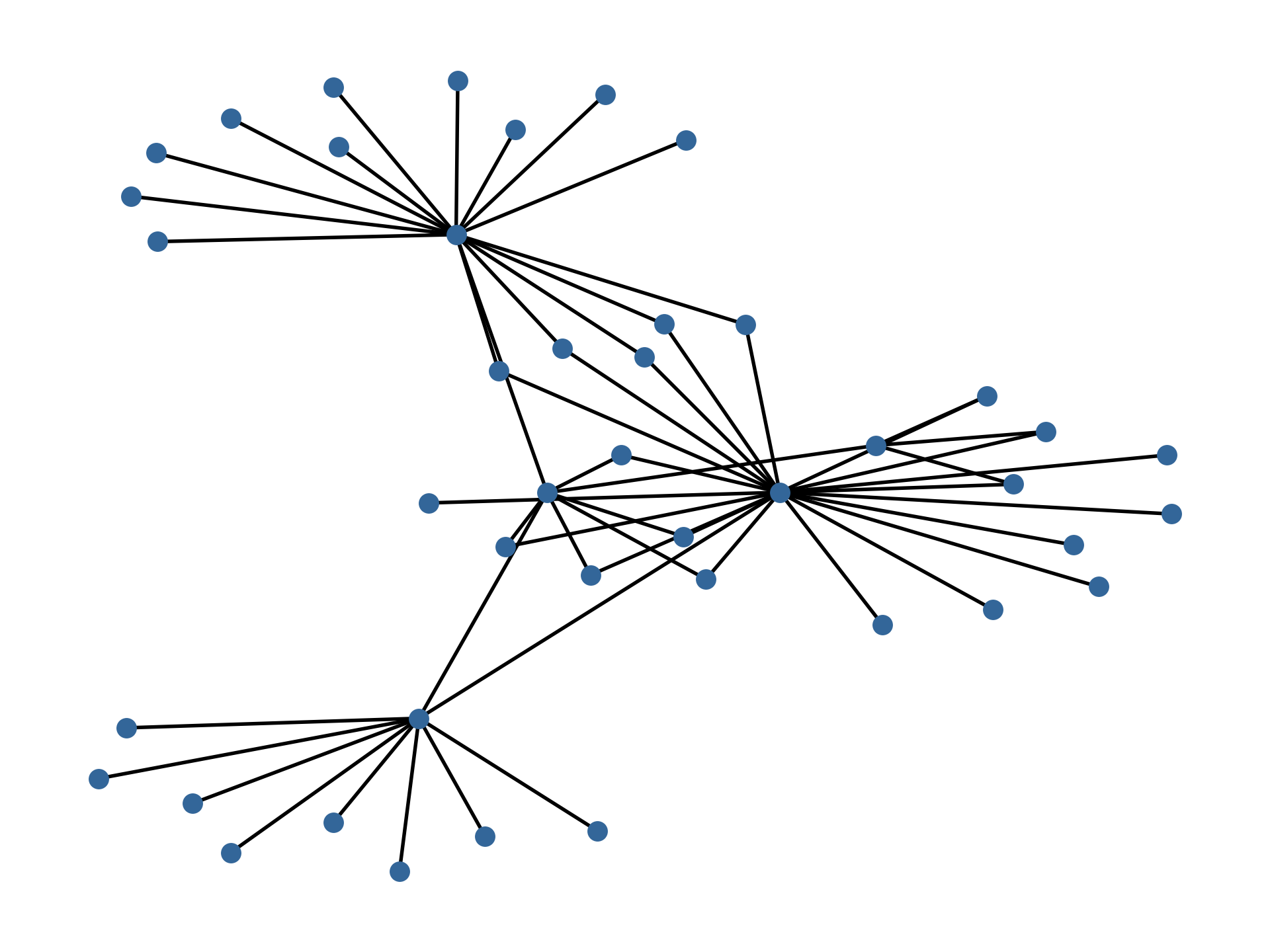}
         &
         \includegraphics[scale=\scl]{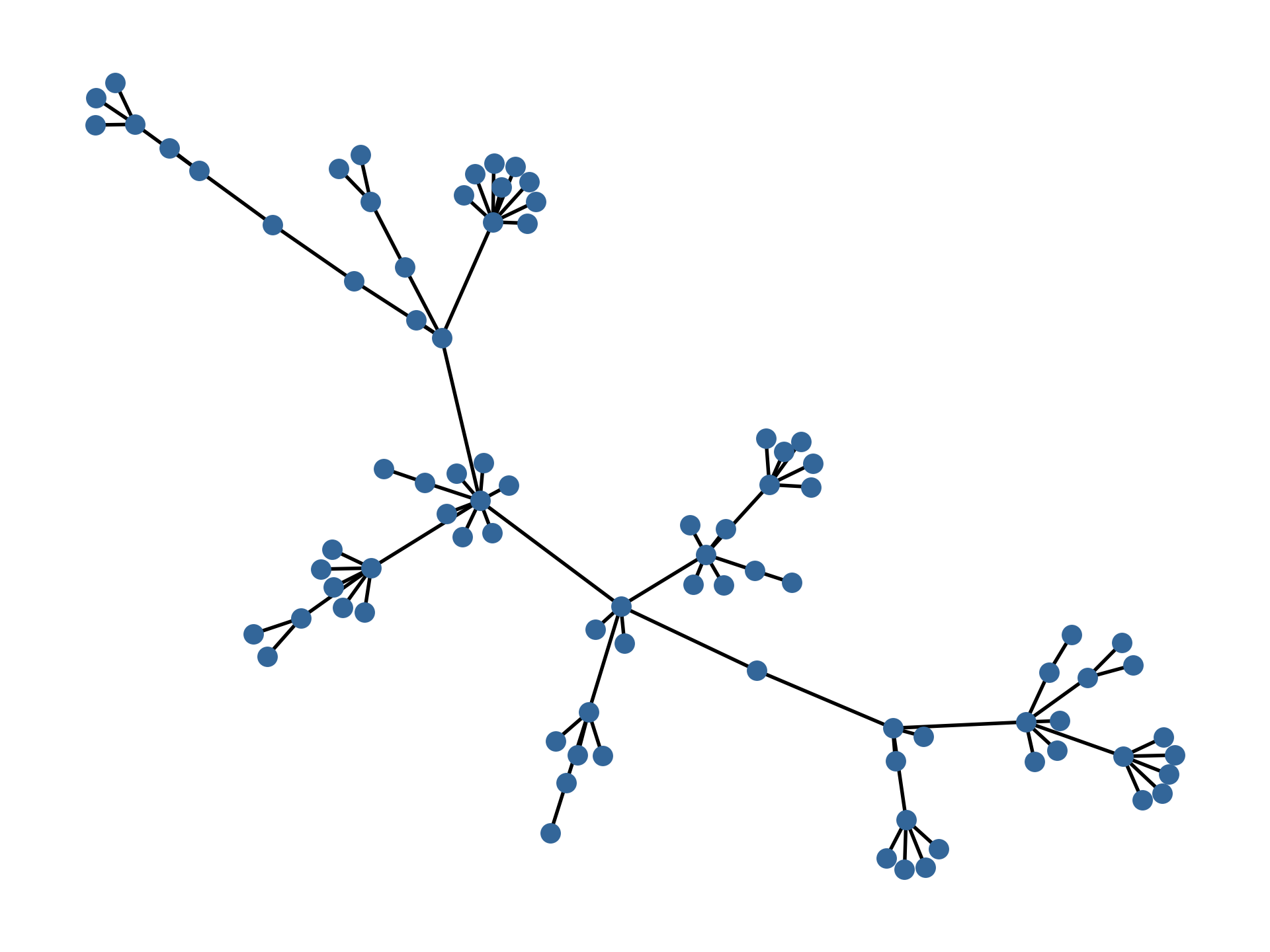}
         &
         \includegraphics[scale=\scl]{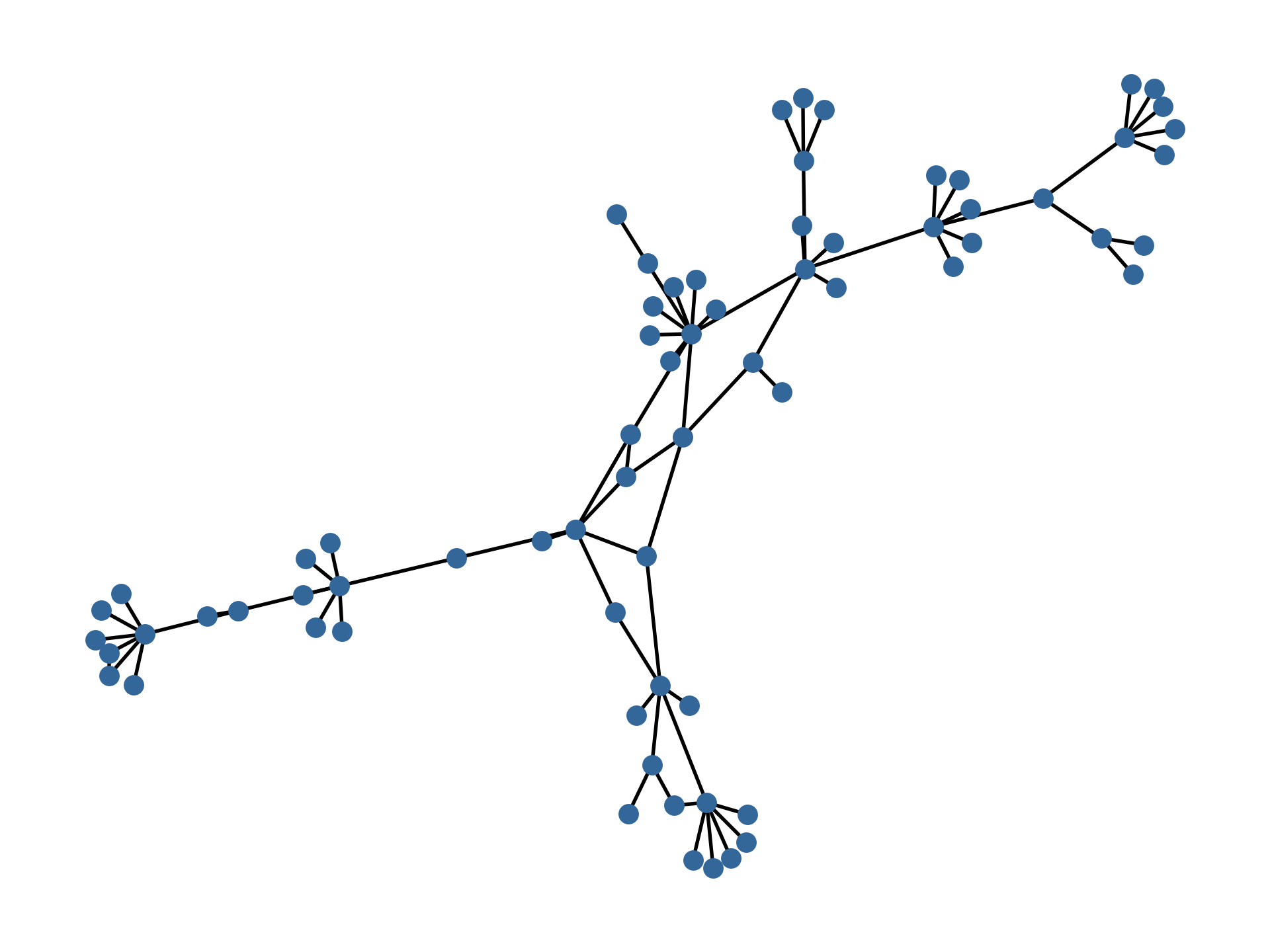}
         \\
         \rotatebox{90}{Lobster}
         & \includegraphics[scale=\scl]{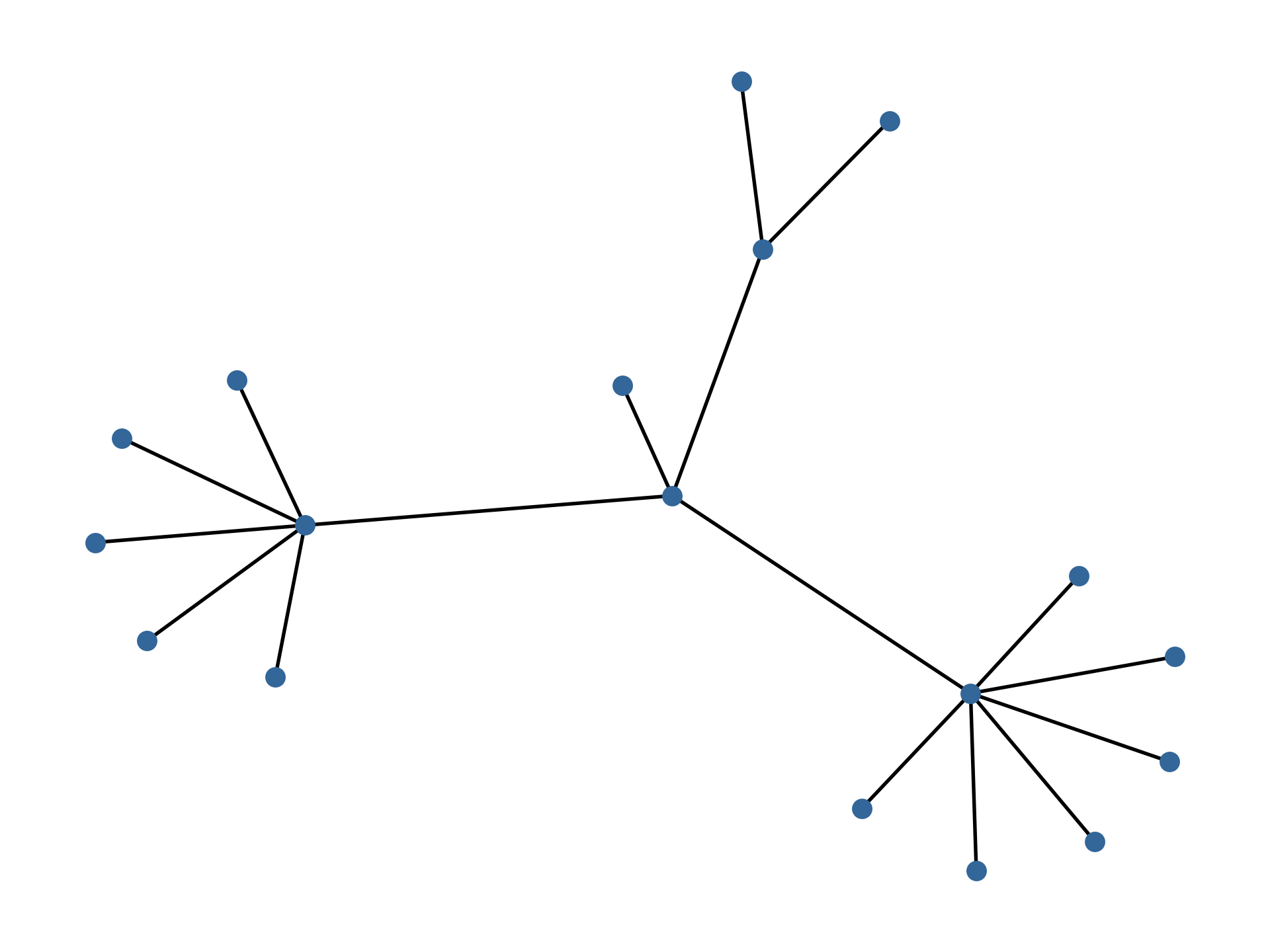} 
         &  \includegraphics[scale=\scl]{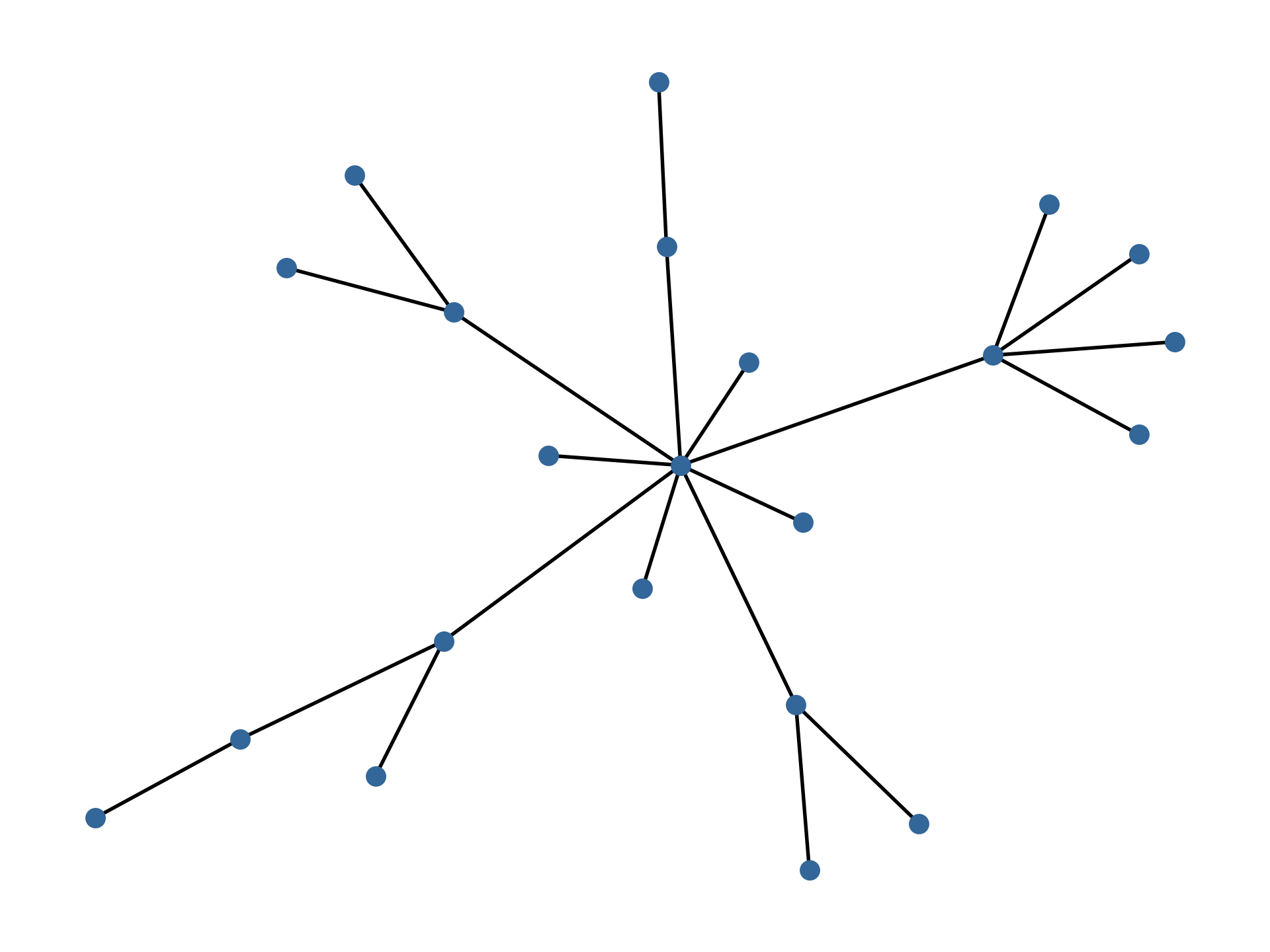} 
         & \includegraphics[scale=\scl]{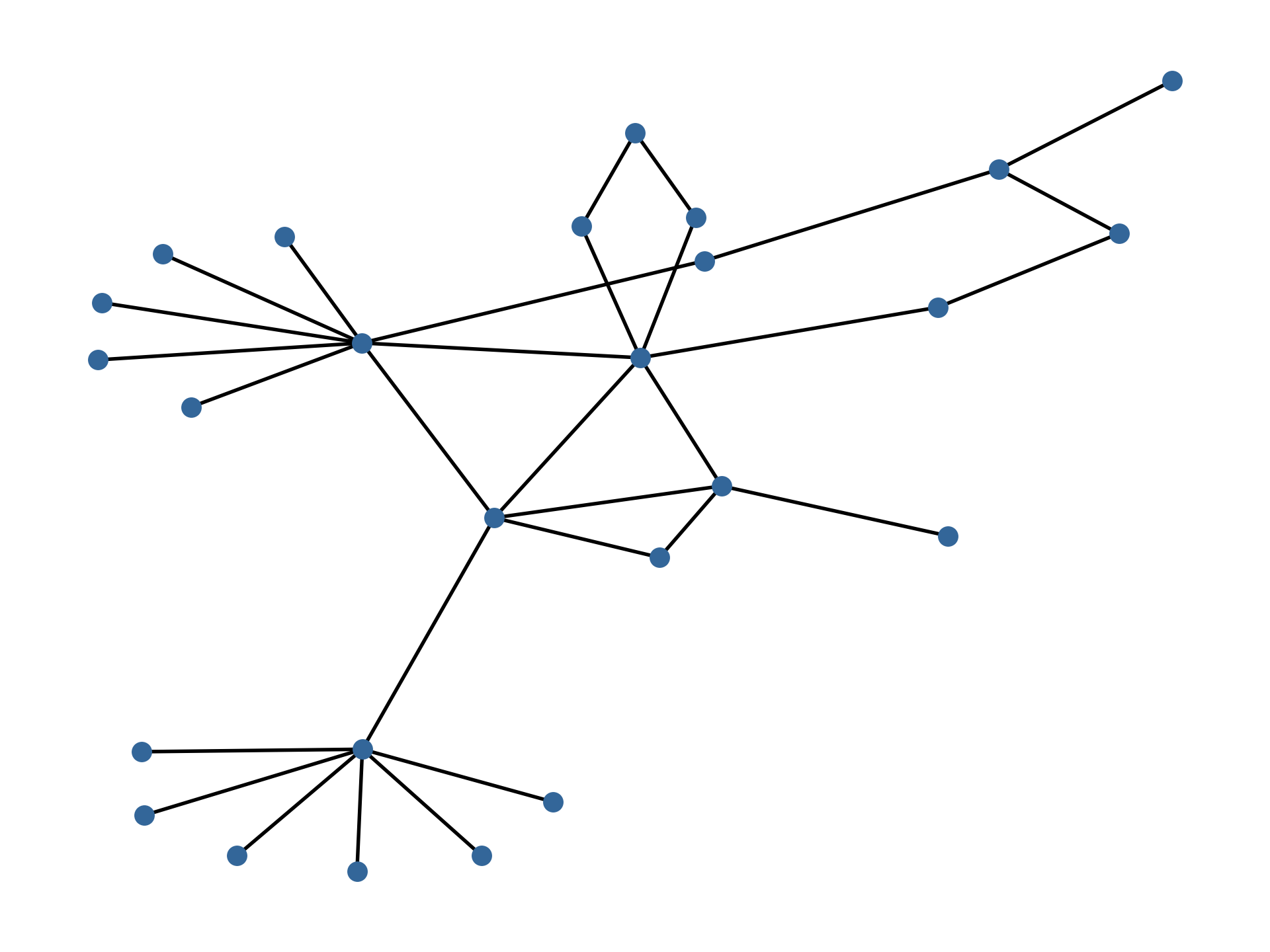}
         &
         \includegraphics[scale=\scl]{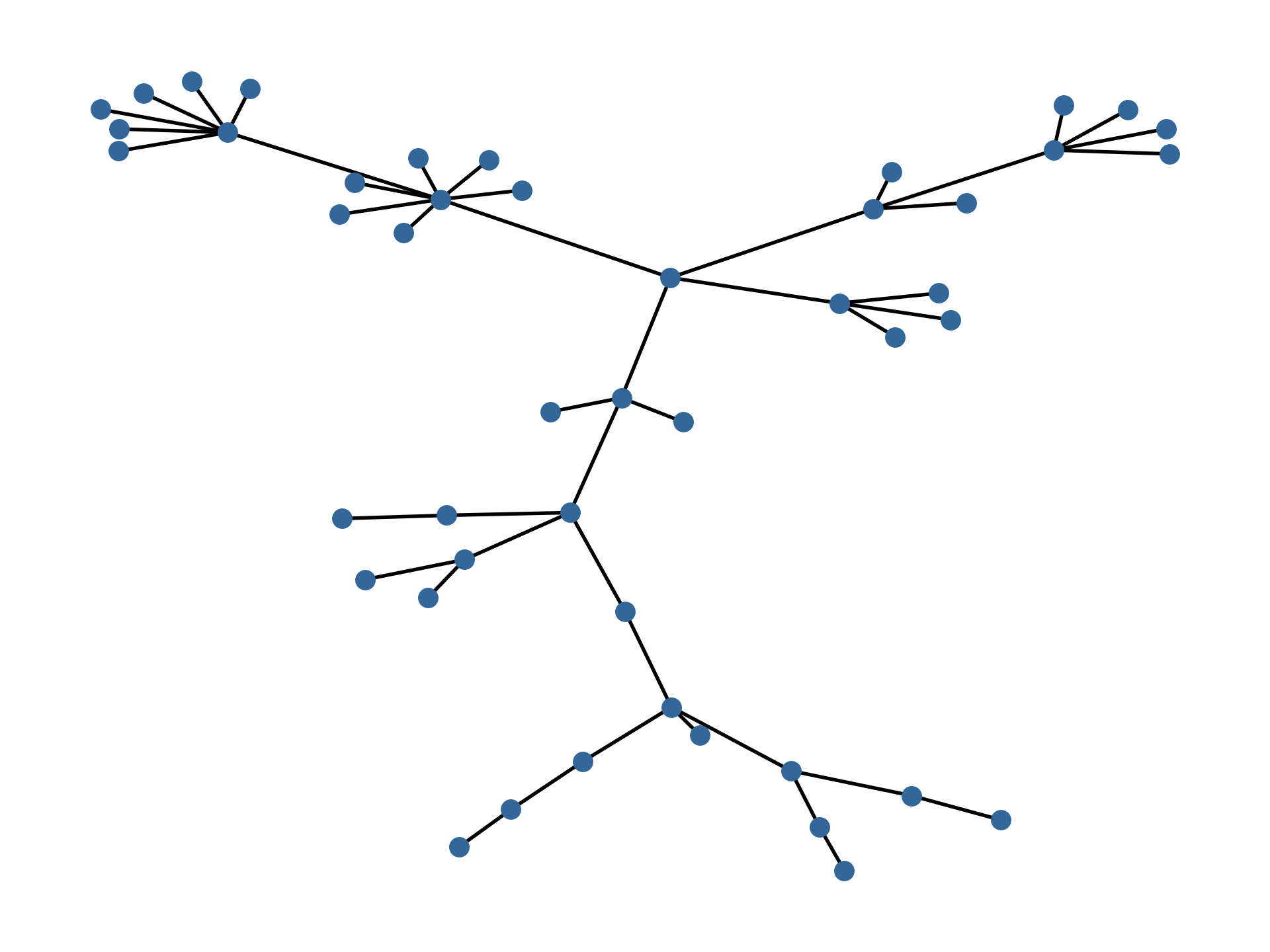}
         &
         \includegraphics[scale=\scl]{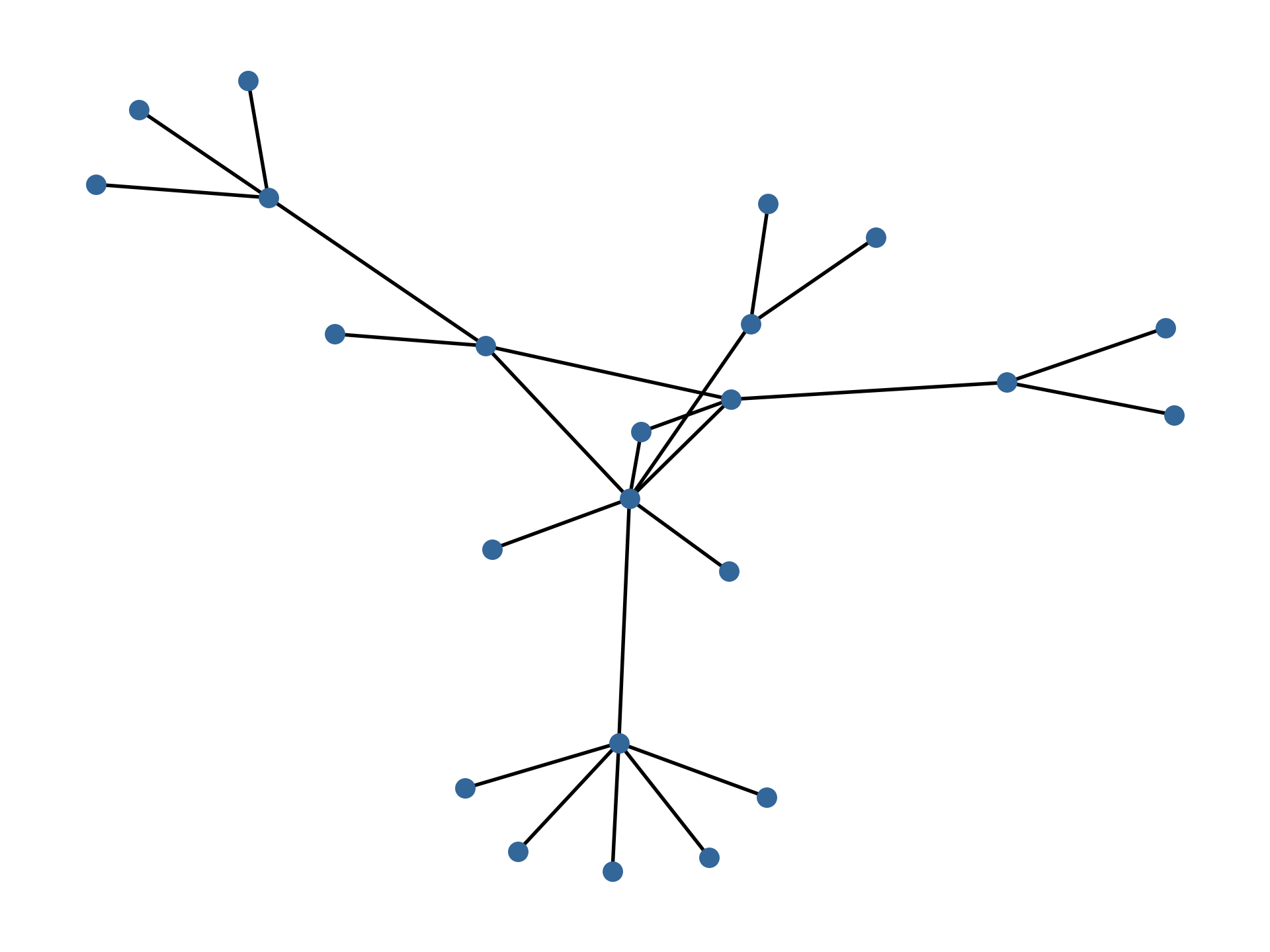}
         \\
    \end{tabular}
    
    \caption{Some original samples and generated graphs by different methods, for each dataset.}
    \label{fig:generatedGraphs}
\end{figure*}

\begin{table}[htbp]
\caption{Ablation study: improvements caused by different parts of Gransformer - Ego dataset}
\label{table:ablation}
\centering
\resizebox{\linewidth}{!} {
\rowcolors{2}{gray!25}{white}
\begin{tabular}{l c  c  c  c  c  c} 
  & using & using & pres/abs& graph positional& Train&Test \\
  & MADE & $K$ & of edges & encoding & NLL & NLL \\
  \midrule
  1 & & & & & 892.71 & 1026.75\\
  2 & & & & \checkmark & 869.07 & 1006.02 \\
  3 & & & \checkmark & \checkmark & 883.35 & 1021.20\\
  4 & & \checkmark & & \checkmark & 849.57 & 980.16 \\
  5 & & \checkmark & \checkmark & & 861.90 & 988.47 \\
  6 & & \checkmark & \checkmark & \checkmark & 852.71 & 990.45 \\
  7 & \checkmark &  &  & \checkmark & 806.97 & 931.77 \\
  8 & \checkmark & \checkmark &  & \checkmark & 794.36 & 916.51 \\
  9 & \checkmark & \checkmark & \checkmark & \checkmark & 790.61 & 909.12 \\
  \bottomrule
\end{tabular}
}
\end{table}

In order to evaluate the proposed Gransformer, we conduct computer experiments.
This section presents our experiments to compare Gransformer with state-of-the-art methods. The implementation of the proposed model and the datasets used in our experiments are available anonymously from \url{https://github.com/Gransformer/Gransformer}.

\subsection{Datasets}
We run the experiments on the following datasets: a synthetic and two real-world. We used 20\% of each dataset as test data and the other 80\% as the training set. For these datasets, we used models with six attention layers ($T=6$) with a hidden size of $400$.\\
   \noindent{\bf Ego:} A real-world dataset, containing 757 graphs representing documents (as nodes) and their citations (as edges) extracted from the Citeseer network \cite{sen2008collective}. Graphs have from 50 up to 399 nodes.\\
   \noindent{\bf Protein:}  A real-world dataset of 918 protein graphs \cite{dobson2003distinguishing}. Nodes are amino acids and geometrically close nodes are connected in the graph. Graph sizes vary from 100 to 500.\\
    \noindent{\bf Lobster:} A synthetic dataset of 100 lobsters of at most 100 nodes (average number of nodes equals 80). A lobster is a tree graph that reduces to a caterpillar if we eliminate all its leaf nodes. A caterpillar is a tree that reduces to a path if we eliminate all its leaf nodes. We used parameters $p_1=0.7$ and $p_2=0.7$ for adding nodes of level $1$ and $2$ to a path to generate a lobster.

\subsection{Evaluation Metrics}
To compare the generated set of graphs with the actual test data, You et al. \cite{graphRNN} suggested using the \emph{maximum mean discrepancy} (MMD) of several graph statistics to compare all moments of the empirical distribution of the graphs. Similar to \cite{graphRNN}, we report MMD score for degree and cluster coefficient distributions, as well as average orbit, counts statistics.
In addition, to study the effect of different ideas proposed in this paper, we report training and test the negative log-likelihood of Gransformer, and Gransformer without some of its modules.

\subsection{Comparison With Other Generative Methods}
In this section, we compare the statistical performance of Gransformer with GraphRNN, GraphRNN-S \cite{graphRNN}, and GRAN \cite{GRAN} as three baseline autoregressive graph generation models. GraphRNN and GraphRNN-S are based on recurrent neural networks, GRAN is designed based on graph convolutional networks, and Gransformer is transformer-based. 
Table \ref{table:MMD} presents the results of different methods. It reports the mean and the standard deviation of the MMD scores for the graphs generated by each method after 2600, 2800, and 3000 training epochs. In each column, the best result is represented in bold digits, and the best two cells are shaded. Table \ref{table:MMD} shows that in most of the columns Gransformer is among the two best results. Moreover, Gransformer is the only method that does not achieve the worst MMD value in any of the columns.

Moreover, Table \ref{table:MMD} shows that neither GraphRNN nor GraphRNN-S outperforms the other one by achieving lower MMD values in all cases. There are some cases where their MMD values differ by an order of magnitude, e.g., the MMD of their clustering coefficients for Ego, or the MMD of their degree distributions for Protein.

From the reported results on Lobster dataset in Table \ref{table:MMD}, it is clear that both graphRNN and Gransformer have detected that there should be no closed triple nodes in the graph, while graphRNN-S and GRAN have been unable to detect this rule from the training graphs.
Figure \ref{fig:generatedGraphs} illustrates several original samples and generated graphs by different methods, for each dataset.

\subsection{Ablation Study} \label{sec:ablation}
We ablate different components of the proposed method to investigate the impact of each one. Table \ref{table:ablation} reports the negative log-likelihood of the ablated models on both train and test data. The columns "using MADE", "using K", "pres/abs of edges" and "graph positional encoding" respectively correspond to the components described in Section \ref{sec:MADE}, Equation \ref{eq:graph-kernel}, Section \ref{sec:typedEdges}, and Equation \ref{eq:graph-pos-enc}.

Comparing row 2 with row 1, it is clear that using graph positional encoding causes a significant improvement in our model. Although our proposed graph positional encoding is computed in a similar way to how the graph-based familiarity matrix $K$ is computed, comparing row 6 with row 5 reveals that even when using $K$, the proposed graph positional encoding causes some improvement in the results.

Comparison between row 3 and row 2, and also between row 6 and row 6 indicates that using presence and absence of edges, as described in Section \ref{sec:typedEdges}, does not necessarily have any gain in practice. It just duplicates the complexity of the attention layers and may even worsen the results. 
This observation emphasizes the necessity of using higher-order proximity of nodes in the graph to extract its structural information. This is exactly what our proposed graph-based familiarity measure $K$ and our proposed graph-based positional encoding were defined based on, in Equations \ref{eq:graph-kernel} and \ref{eq:graph-pos-enc}

The significant improvement in NLL values in row 7 against all rows 1 to 6 confirms the significant impact of dependent edge generation using MADE in Gransformer. Also, comparing row 7 with rows 8 and 9 verifies that other components are still advantageous in the presence of MADE.

Finally, since the results in rows 8 and 9 are very similar, we keep using the idea of utilizing presence and absence of edges (Section \ref{sec:typedEdges}) in our model, although it does not necessarily have a significant impact on the performance of the model.


\section{Conclusion} \label{sec:conclusion}
In this paper, we proposed Gransformer, an autoregressive graph generation model based on a single transformer encoder. We extended the transformer encoder using several techniques. We put an autoregressive mask on the encoder and modified the attention mechanism to utilize the presence or absence of edges between nodes. 
We also imported graph structural properties to the model by defining a familiarity measure between node pairs inspired by message-passing algorithms. We used the proposed familiarity measure both in the attention layer and for graph-based positional encoding. Our experiments showed that our method is competitive with other methods respecting the MMD scores.
	\bibliographystyle{IEEEtran}
    \bibliography{IEEEabrv,ref}

\begin{thebibliography}{10}
\providecommand{\url}[1]{#1}
\csname url@samestyle\endcsname
\providecommand{\newblock}{\relax}
\providecommand{\bibinfo}[2]{#2}
\providecommand{\BIBentrySTDinterwordspacing}{\spaceskip=0pt\relax}
\providecommand{\BIBentryALTinterwordstretchfactor}{4}
\providecommand{\BIBentryALTinterwordspacing}{\spaceskip=\fontdimen2\font plus
\BIBentryALTinterwordstretchfactor\fontdimen3\font minus \fontdimen4\font\relax}
\providecommand{\BIBforeignlanguage}[2]{{%
\expandafter\ifx\csname l@#1\endcsname\relax
\typeout{** WARNING: IEEEtran.bst: No hyphenation pattern has been}%
\typeout{** loaded for the language `#1'. Using the pattern for}%
\typeout{** the default language instead.}%
\else
\language=\csname l@#1\endcsname
\fi
#2}}
\providecommand{\BIBdecl}{\relax}
\BIBdecl

\bibitem{graphRNN}
J.~You, R.~Ying, X.~Ren, W.~Hamilton, and J.~Leskovec, ``Graphrnn: Generating realistic graphs with deep auto-regressive models,'' in \emph{Proceedings of International Conference on Machine Learning}, 2018, pp. 5708--5717.

\bibitem{GRAN}
R.~Liao, Y.~Li, Y.~Song, S.~Wang, W.~Hamilton, D.~K. Duvenaud, R.~Urtasun, and R.~Zemel, ``Efficient graph generation with graph recurrent attention networks,'' \emph{Advances in Neural Information Processing Systems}, vol.~32, pp. 4255--4265, 2019.

\bibitem{GRAM}
W.~Kawai, Y.~Mukuta, and T.~Harada, ``Scalable generative models for graphs with graph attention mechanism,'' \emph{arXiv preprint arXiv:1906.01861}, 2019.

\bibitem{DeepGMG}
Y.~Li, O.~Vinyals, C.~Dyer, R.~Pascanu, and P.~Battaglia, ``Learning deep generative models of graphs,'' \emph{arXiv preprint arXiv:1803.03324}, 2018.

\bibitem{AGE}
S.~Fan and B.~Huang, ``Attention-based graph evolution,'' \emph{Advances in Knowledge Discovery and Data Mining}, vol. 12084, p. 436, 2020.

\bibitem{MADE}
M.~Germain, K.~Gregor, I.~Murray, and H.~Larochelle, ``Made: Masked autoencoder for distribution estimation,'' in \emph{Proceedings of International Conference on Machine Learning}, 2015, pp. 881--889.

\bibitem{papamakarios2017masked}
G.~Papamakarios, T.~Pavlakou, and I.~Murray, ``Masked autoregressive flow for density estimation,'' \emph{arXiv preprint arXiv:1705.07057}, 2017.

\bibitem{khajenezhad2020masked}
A.~Khajenezhad, H.~Madani, and H.~Beigy, ``Masked autoencoder for distribution estimation on small structured data sets,'' \emph{IEEE Transactions on Neural Networks and Learning Systems}, vol.~32, no.~11, pp. 4997--5007, 2021.

\bibitem{faez2021deep}
F.~Faez, Y.~Ommi, M.~S. Baghshah, and H.~R. Rabiee, ``Deep graph generators: A survey,'' \emph{IEEE Access}, vol.~9, pp. 106\,675--106\,702, 2021.

\bibitem{BiGG}
H.~Dai, A.~Nazi, Y.~Li, B.~Dai, and D.~Schuurmans, ``Scalable deep generative modeling for sparse graphs,'' in \emph{Proceedings of International Conference on Machine Learning}, 2020, pp. 2302--2312.

\bibitem{VGAE}
T.~N. Kipf and M.~Welling, ``Variational graph auto-encoders,'' \emph{Advances in Neural Information Processing Systems: Workshop on Bayesian Deep Learning}, 2016.

\bibitem{GraphVAE}
M.~Simonovsky and N.~Komodakis, ``Graphvae: Towards generation of small graphs using variational autoencoders,'' in \emph{Proceedings of International conference on artificial neural networks}, 2018, pp. 412--422.

\bibitem{Nevae}
B.~Samanta, A.~De, G.~Jana, V.~G{\'o}mez, P.~K. Chattaraj, N.~Ganguly, and M.~Gomez-Rodriguez, ``Nevae: A deep generative model for molecular graphs,'' \emph{Journal of machine learning research}, vol.~21, no. 114, pp. 1--33, April 2020.

\bibitem{Netgan}
A.~Bojchevski, O.~Shchur, D.~Z{\"u}gner, and S.~G{\"u}nnemann, ``Netgan: Generating graphs via random walks,'' in \emph{Proceedings of International Conference on Machine Learning}, 2018, pp. 610--619.

\bibitem{condgen}
C.~Yang, P.~Zhuang, W.~Shi, A.~Luu, and P.~Li, ``Conditional structure generation through graph variational generative adversarial nets.'' in \emph{Advances in Neural Information Processing Systems}, 2019, pp. 1338--1349.

\bibitem{molgan}
N.~De~Cao and T.~Kipf, ``{MolGAN: An implicit generative model for small molecular graphs},'' in \emph{Proceedings of International Conference on Machine Learning: Workshop on Theoretical Foundations and Applications of Deep Generative Models}, 2018.

\bibitem{gcpn}
J.~You, B.~Liu, R.~Ying, V.~Pande, and J.~Leskovec, ``Graph convolutional policy network for goal-directed molecular graph generation,'' \emph{Advances in Neural Information Processing Systems}, 2018.

\bibitem{graphopt}
R.~Trivedi, J.~Yang, and H.~Zha, ``Graphopt: Learning optimization models of graph formation,'' in \emph{Proceedings of International Conference on Machine Learning}, 2020, pp. 9603--9613.

\bibitem{liu2019graph}
J.~Liu, A.~Kumar, J.~Ba, J.~Kiros, and K.~Swersky, ``Graph normalizing flows,'' \emph{arXiv preprint arXiv:1905.13177}, 2019.

\bibitem{madhawa2019graphnvp}
K.~Madhawa, K.~Ishiguro, K.~Nakago, and M.~Abe, ``Graphnvp: An invertible flow model for generating molecular graphs,'' \emph{arXiv preprint arXiv:1905.11600}, 2019.

\bibitem{shi2020graphaf}
C.~Shi, M.~Xu, Z.~Zhu, W.~Zhang, M.~Zhang, and J.~Tang, ``Graphaf: a flow-based autoregressive model for molecular graph generation,'' \emph{arXiv preprint arXiv:2001.09382}, 2020.

\bibitem{diff1}
C.~Niu, Y.~Song, J.~Song, S.~Zhao, A.~Grover, and S.~Ermon, ``Permutation invariant graph generation via score-based generative modeling,'' in \emph{Proceedings of International Conference on Artificial Intelligence and Statistics}, 2020, pp. 4474--4484.

\bibitem{diff2}
X.~Chen, Y.~Li, A.~Zhang, and L.-p. Liu, ``Nvdiff: Graph generation through the diffusion of node vectors,'' \emph{arXiv preprint arXiv:2211.10794}, 2022.

\bibitem{diff3}
H.~Huang, L.~Sun, B.~Du, Y.~Fu, and W.~Lv, ``Graphgdp: Generative diffusion processes for permutation invariant graph generation,'' in \emph{Proceedings of IEEE International Conference on Data Mining (ICDM)}, 2022, pp. 201--210.

\bibitem{diff4}
J.~Jo, S.~Lee, and S.~J. Hwang, ``Score-based generative modeling of graphs via the system of stochastic differential equations,'' in \emph{Proceedings of International Conference on Machine Learning}, 2022, pp. 10\,362--10\,383.

\bibitem{diff5}
T.~Luo, Z.~Mo, and S.~J. Pan, ``Fast graph generative model via spectral diffusion,'' \emph{arXiv preprint arXiv:2211.08892}, 2022.

\bibitem{MPGVAE}
D.~Flam-Shepherd, T.~Wu, and A.~Aspuru-Guzik, ``Graph deconvolutional generation,'' \emph{arXiv preprint arXiv:2002.07087}, 2020.

\bibitem{MolMP}
Y.~Li, L.~Zhang, and Z.~Liu, ``Multi-objective de novo drug design with conditional graph generative model,'' \emph{Journal of cheminformatics}, vol.~10, no.~1, pp. 1--24, 2018.

\bibitem{MolecularRNN}
M.~Popova, M.~Shvets, J.~Oliva, and O.~Isayev, ``Molecularrnn: Generating realistic molecular graphs with optimized properties,'' \emph{arXiv preprint arXiv:1905.13372}, 2019.

\bibitem{bacciu2019graph}
D.~Bacciu, A.~Micheli, and M.~Podda, ``Graph generation by sequential edge prediction,'' in \emph{Proceedings of the European Symposium on Artificial Neural Networks, Computational Intelligence and Machine Learning}, 2019, pp. 95--100.

\bibitem{bacciu2020edge}
------, ``Edge-based sequential graph generation with recurrent neural networks,'' \emph{Neurocomputing}, vol. 416, pp. 177--189, 2020.

\bibitem{GraphGen}
N.~Goyal, H.~V. Jain, and S.~Ranu, ``Graphgen: a scalable approach to domain-agnostic labeled graph generation,'' in \emph{Proceedings of The Web Conference}, 2020, pp. 1253--1263.

\bibitem{GrAD}
S.~A. Shah and V.~Koltun, ``Auto-decoding graphs,'' \emph{arXiv preprint arXiv:2006.02879}, 2020.

\bibitem{MMGAN}
A.~Gamage, E.~Chien, J.~Peng, and O.~Milenkovic, ``Multi-motifgan (mmgan): Motif-targeted graph generation and prediction,'' in \emph{Proceedings of the IEEE International Conference on Acoustics, Speech and Signal Processing (ICASSP)}, 2020, pp. 4182--4186.

\bibitem{DGVAE}
J.~Li, J.~Yu, J.~Li, H.~Zhang, K.~Zhao, Y.~Rong, H.~Cheng, and J.~Huang, ``Dirichlet graph variational autoencoder,'' \emph{Advances in Neural Information Processing Systems}, vol.~33, pp. 5274--5283, 2020.

\bibitem{Graphite}
A.~Grover, A.~Zweig, and S.~Ermon, ``Graphite: Iterative generative modeling of graphs,'' in \emph{Proceedings of International Conference on Machine Learning}, 2019, pp. 2434--2444.

\bibitem{DEFactor}
R.~Assouel, M.~Ahmed, M.~H. Segler, A.~Saffari, and Y.~Bengio, ``Defactor: Differentiable edge factorization-based probabilistic graph generation,'' \emph{arXiv preprint arXiv:1811.09766}, 2018.

\bibitem{DeepWalk}
B.~Perozzi, R.~Al-Rfou, and S.~Skiena, ``Deepwalk: Online learning of social representations,'' in \emph{Proceedings of the ACM SIGKDD international conference on Knowledge discovery and data mining}, 2014, pp. 701--710.

\bibitem{Node2Vec}
A.~Grover and J.~Leskovec, ``node2vec: Scalable feature learning for networks,'' in \emph{Proceedings of the ACM SIGKDD international conference on Knowledge discovery and data mining}, 2016, pp. 855--864.

\bibitem{SimNet}
M.~Khajehnejad, ``{SimNet}: Similarity-based network embeddings with mean commute time,'' \emph{PloS one}, vol.~14, no.~8, p. e0221172, 2019.

\bibitem{GCNClassification}
T.~N. Kipf and M.~Welling, ``Semi-supervised classification with graph convolutional networks,'' in \emph{Proceedings of the International Conference on Learning Representations}, 2017.

\bibitem{GraphSage}
W.~Hamilton, Z.~Ying, and J.~Leskovec, ``Inductive representation learning on large graphs,'' \emph{Advances in neural information processing systems}, vol.~30, 2017.

\bibitem{bert}
J.~Devlin, M.-W. Chang, K.~Lee, and K.~Toutanova, ``Bert: Pre-training of deep bidirectional transformers for language understanding,'' \emph{arXiv preprint arXiv:1810.04805}, 2018.

\bibitem{gpt3}
T.~Brown, B.~Mann, N.~Ryder, M.~Subbiah, J.~D. Kaplan, P.~Dhariwal, A.~Neelakantan, P.~Shyam, G.~Sastry, A.~Askell, S.~Agarwal, A.~Herbert-Voss, G.~Krueger, T.~Henighan, R.~Child, A.~Ramesh, D.~Ziegler, J.~Wu, C.~Winter, C.~Hesse, M.~Chen, E.~Sigler, M.~Litwin, S.~Gray, B.~Chess, J.~Clark, C.~Berner, S.~McCandlish, A.~Radford, I.~Sutskever, and D.~Amodei, ``Language models are few-shot learners,'' in \emph{Advances in Neural Information Processing Systems}, vol.~33, 2020, pp. 1877--1901.

\bibitem{transformers}
A.~Vaswani, N.~Shazeer, N.~Parmar, J.~Uszkoreit, L.~Jones, A.~N. Gomez, {\L}.~Kaiser, and I.~Polosukhin, ``Attention is all you need,'' in \emph{Advances in Neural Information Processing Systems}, 2017, pp. 5998--6008.

\bibitem{matrixMult}
J.~Alman and V.~V. Williams, ``A refined laser method and faster matrix multiplication,'' in \emph{Proceedings of the ACM-SIAM Symposium on Discrete Algorithms (SODA)}, 2021, pp. 522--539.

\bibitem{strassen1969gaussian}
V.~Strassen, ``Gaussian elimination is not optimal,'' \emph{Numerische mathematik}, vol.~13, no.~4, pp. 354--356, 1969.

\bibitem{sen2008collective}
P.~Sen, G.~Namata, M.~Bilgic, L.~Getoor, B.~Galligher, and T.~Eliassi-Rad, ``Collective classification in network data,'' \emph{AI magazine}, vol.~29, no.~3, pp. 93--93, 2008.

\bibitem{dobson2003distinguishing}
P.~D. Dobson and A.~J. Doig, ``Distinguishing enzyme structures from non-enzymes without alignments,'' \emph{Journal of molecular biology}, vol. 330, no.~4, pp. 771--783, 2003.

\end{thebibliography}

\end{document}